\documentclass[letterpaper, 10 pt, conference]{ieeeconf}

\IEEEoverridecommandlockouts                              
\usepackage{tikz}
\newcommand\submittedtext{%
  \footnotesize This work has been published in IEEE ACCES \url{https://doi.org/10.1109/ACCESS.2025.3596857}}

\newcommand\submittednotice{%
\begin{tikzpicture}[remember picture,overlay]
\node[anchor=south,yshift=10pt] at (current page.south) {\fbox{\parbox{\dimexpr0.65\textwidth-\fboxsep-\fboxrule\relax}{\submittedtext}}};
\end{tikzpicture}%
}

\usepackage{amsmath}

\usepackage{hyperref}
\usepackage{subcaption}
\usepackage{makecell}
\title{\LARGE \bf Detection and Tracking of MAVs Using a\\ Rosette Scanning Pattern LiDAR}

\author{Sándor Gazdag$^{1,2}$, Tom Möller$^{1,3,4}$, Anita Keszler$^{1}$ and András L. Majdik$^{1,2}$% <-this % stops a space
\thanks{*
This work was supported in part by AFOSR Award No. FA8655-23-1-7071, by the European Union within the framework of the National Laboratory for Autonomous Systems (RRF-2.3.1-21-2022-00002), by the Hungarian Scientific Research Fund (No. NKFIH OTKA K-139485), by TKP2021-NVA-01 project and by the Doctoral Excellence Fellowship Programme (DCEP) of the NRDI fund and BME under a grant agreement with the NRDI office.}% <-this % stops a space 
\thanks{$^{1}$HUN-REN SZTAKI Institute for Computer Science and Control, Kende u. 13-17, 1111 Budapest, Hungary (e-mail: sandor.gazdag, tom.moller, keszler.anita, andras.majdik\}@sztaki.hun-ren.hu}%        
\thanks{$^{2}$Budapest University of Technology and Economics, Műegyetem rkp. 3. 1111 Budapest, Hungary}
\thanks{$^{3}$Eötvös Loránd University (ELTE), Egyetem tér 1-3, 1053 Budapest, Hungary}
\thanks{$^{4}$KTH Royal Institute of Technology, SE-100 44 Stockholm, Sweden}
}

\usepackage{graphicx}

\usepackage{lineno}
\usepackage{xcolor}
\begin{document}
\maketitle
\submittednotice
\setcounter{footnote}{3}

%%%%%%%%%%%%%%%%%%%%%%%%%%%%%%%%%%%%%%%%%%%%%%%%%%%%%%%%%%%%%%%%%%%%%%%%%%%%%%%%%%%%%
%                                   ABSTRACT
%%%%%%%%%%%%%%%%%%%%%%%%%%%%%%%%%%%%%%%%%%%%%%%%%%%%%%%%%%%%%%%%%%%%%%%%%%%%%%%%%%%%%
\begin{abstract}
The use of commercial Micro Aerial Vehicles (MAVs) has surged in the past decade, offering societal benefits but also raising risks such as airspace violations and privacy concerns. Due to the increased security risks, the development of autonomous drone detection and tracking systems has become a priority. In this study, we tackle this challenge, by using non-repetitive rosette scanning pattern LiDARs, particularly focusing on increasing the detection distance by leveraging the characteristics of the sensor. The presented method utilizes a particle filter with a velocity component for the detection and tracking of the drone, which offers added re-detection capability. A pan-tilt platform is utilized to take advantage of the specific characteristics of the rosette scanning pattern LiDAR by keeping the tracked object in the center where the measurement is most dense. The system's tracking capabilities (both in coverage and distance), as well as its accuracy are validated and compared to State Of The Art (SOTA) models, demonstrating improved performance, particularly in terms of coverage and maximum tracking distance. Our approach achieved accuracy on par with the SOTA indoor method while increasing the maximum detection range by approximately $85\;\%$ beyond the SOTA outdoor method to $130\;m$. Additionally, our method yields at least a twofold increase in track coverage and returned point counts.
\end{abstract}


%%%%%%%%%%%%%%%%%%%%%%%%%%%%%%%%%%%%%%%%%%%%%%%%%%%%%%%%%%%%%%%%%%%%%%%%%%%%%%%%%%%%%
%                               INTRODUCTION
%%%%%%%%%%%%%%%%%%%%%%%%%%%%%%%%%%%%%%%%%%%%%%%%%%%%%%%%%%%%%%%%%%%%%%%%%%%%%%%%%%%%%
% Change 2.\section{Introduction}
\label{introduction}
\PARstart{M}{icro} Air Vehicles (MAVs) have contributed to the development of many fields, such as precision agriculture \cite{Rejeb_2022}, modern production \cite{Maghazei_2019}, transport industry \cite{Chi-2023} and mining industry \cite{Shahmoradi-2020}.
In addition to the above, the utilization of drones (in this paper the phrases drone and MAV are interchangeable) for hobby purposes is also developing explosively, because they have become affordable and easy to handle \cite{Mohsan_2023}, increasing the risk of inappropriate use. Hobby drones can be used for unauthorized recording, violating people's right to undisturbed privacy \cite{Wilson-2014, Vacca-2017}. Furthermore, they can also threaten protected buildings (e.g. airports, factories, power plants) \cite{Lukasiewicz2022}. Due to the increased security risks, the development of autonomous drone detection and tracking systems has become a priority area of research.

%%%%%%%%%%%%%%%% problem definition
%Ennek a cikknek a lényege, hogy megvizsgáljunk milyen távolságban lehet repülő repüló obiektumot detektálni és követni a rozettával.
In this study, we tackle the challenge of drone detection and tracking using non-repetitive rosette scanning pattern LiDARs, particularly focusing on increasing the detection distance by leveraging the characteristics of the sensor.

We use the rosette scanning pattern LIVOX AVIA LiDAR sensor, that uses multiple, aligned laser beams. The beams move in a non-repetitive rosette-like pattern inside the sensor, that covers the entire Field of View (FoV) by using Risley prisms \cite{brazeal2021risley}. This type of sensor design results in faster measurements, and cheaper devices with fewer moving components compared to the more common rotating models. An important feature of this type of LiDAR is that it captures more measurement data from a target object in the center of the sensor’s FoV, as the laser beams pass through the center each time they travel along a rotating 8-shaped pattern.

%%%%%%%%%%%%%%%% SOA - the current state of the art (SOA)
Several sensors can be used for drone detection, such as radar \cite{Coluccia_2020, Zhao_2019}, LiDAR \cite{Hammer_2018, Dogru_2022}, camera \cite{vb01, vb02, vb04}, radio-frequency \cite{Nemer_2021, Medaiyese-2022} and acoustic sensors \cite{Corzo_2023, Al-Emadi_2021}. The detection efficiency can be further improved by using sensor fusion  \cite{svanstrom2020realtime, Dudczyk_2022}. However, each sensor modality should also give good results when used independently \cite{svanstrom2022}.

%%%%%%%%%%%%%%%% limitation of SOA
The State Of The Art (SOTA) drone tracking algorithms using LiDARs either use a classical rotating LiDAR \cite{Dogru_2022} or place a rosette scanning pattern sensor on a stationary mount \cite{catalano_2023}. Dogru et al.\cite{Dogru_2022} achieve a detection distance of $70\;m$ outside using a Velodyne VLP-16\footnote{\href{https://velodynelidar.com/products/puck/}{https://velodynelidar.com/products/puck/}}. This detection distance can only be extended to $100\;m$ by adding a retroreflector to the drone. Catalano et al.\cite{catalano_2023} use a LIVOX Horizon\footnote{\href{https://www.livoxtech.com/3296f540ecf5458a8829e01cf429798e/assets/horizon/LIVOX\%20Horizon\%20user\%20manual\%20v1.0.pdf}{https://www.livoxtech.com/}} to achieve a few $cm$ accuracy using a method based on Kalman Filter which has no re-detection capability in case of lost tracking. Their method was only tested indoors so there is no information on the maximum detection distance.

%%%%%%%%%%%%%%%% beyond SOA - why is our method/algorithm beyond SOA
Our proposed algorithm uses a Particle Filter (PF) with a velocity component to achieve fast detection and tracking with added re-detection capability. Leveraging the point distribution of the LIVOX AVIA LiDAR\footnote{\href{https://www.livoxtech.com/avia}{https://www.livoxtech.com/avia}}, the proposed method keeps the MAV in the middle of the sensor's FoV, where the measurements are most dense, using a pan-tilt platform. With this approach we achieved accuracy on par with the SOTA indoor method using a similar sensor \cite{catalano_2023}, while increasing the maximum detection range approximately $85\%$ beyond the SOTA outdoor method \cite{Dogru_2022} by detecting drones up to $130\;m$ away without attached retroreflector. Additionally, our method yields significantly higher track coverage, meaning the drone is tracked for a greater percentage of its trajectory. Furthermore, the number of points reflected from the drone at the same distance increased several folds over the full range. We provide detailed experiments and comparisons to support these claims and we also analyze how much each component of our algorithm contributes to the observed improvement in tracking distance.

%%%%%%%%%%%%%%%% assumptionokat 
To focus on this challenge, we made two assumptions: (i) the background of the detection area is assumed to be static, and (ii) we disregard the classification problem and assume that all moving objects in the area are drones.

%%%%%%%%%%%%%%%% contributions - contributions in general with respect to SOA
To summarize, this paper enhances the SOTA in LiDAR-based drone tracking with the following contributions:
\begin{itemize}
    \item A novel MAV tracking approach for LiDAR utilizing a particle filter with a velocity component is proposed. The updating of particle states of the filter is straightforward using the LiDAR point clouds and the proposed solution offers added re-detection capabilities.
    \item Our method uses a pan-tilt platform to take advantage of the specific characteristics of the rosette scanning pattern LiDAR by keeping the tracked object in the center where the measurement is most dense.
    \item The system's tracking capabilities (both in coverage and distance), as well as its accuracy are validated and compared to SOTA models \cite{Dogru_2022, catalano_2023}, demonstrating improved performance, particularly in terms of coverage and maximum tracking distance. A video is shared of the experiments.\footnote{\label{footnote:vid}Video demonstration of experiments at \href{https://youtu.be/ghfjqAnDuag}{https://youtu.be/ghfjqAnDuag}}
    \item We release the source code of the proposed algorithm and the datasets used in the experiments.\footnote{\label{footnote:git}Source code at \url{https://github.com/gsanya/LIVOX_MAV_track}}
\end{itemize}

%%%%%%%%%%%%%%%% remainder 
The remainder of the paper is organized as follows: related work is presented in Section \ref{related}, the proposed algorithm in Section \ref{proposed_algorithm}, the experiments in Section \ref{experiments}, and finally the conclusion in Section \ref{conclusion}.

%%%%%%%%%%%%%%%%%%%%%%%%%%%%%%%%%%%%%%%%%%%%%%%%%%%%%%%%%%%%%%%%%%%%%%%%%%%%%%%%%%%%%
%                               RELATED WORK
%%%%%%%%%%%%%%%%%%%%%%%%%%%%%%%%%%%%%%%%%%%%%%%%%%%%%%%%%%%%%%%%%%%%%%%%%%%%%%%%%%%%%

\section{RELATED WORK} \label{related}
Tracking MAVs is challenging due to their small size and rapid movement. Traditionally, vision-based approaches have dominated this field. Methods leveraging neural networks, specifically YOLO \cite{yolo2016}, are widely used for detecting and classifying objects in images and videos. These methods are highly effective in controlled environments with good visibility, as shown in works like \cite{vb01, vb02, vb04, MIMO_new_2022}. Vision-based techniques leverage large datasets and complex neural network architectures to improve accuracy, but they often struggle in outdoor scenarios where lighting and occlusions affect visibility. Furthermore, the maximum detection distance is not only limited by the resolution of the camera but also the input resolution of the neural network used. In a recent work \cite{hammer2020imagebased}, the maximum detection range of such a system is 40 m.

Other than detection, recent works have also focused on visual tracking. The rapid development of deep learning-based visual tracking methods is explored in \cite{Marvasti2022}. Learned motion models have been shown to outperform traditional Kalman Filter-based approaches, particularly in handling non-linear motion \cite{Adzemovic2025}. However, these methods typically rely on dense visual input and are not well-suited for sparse point clouds with a limited number of measurement points. Moreover, these works generally target generic object tracking rather than specifically addressing the challenges of tracking small, fast-moving drones.

In response to these limitations, there is a growing interest in non-camera-based solutions, particularly LiDAR, which provides accurate 3D spatial information. LiDAR offers a distinct advantage in detecting and tracking MAVs in low-visibility conditions, where cameras are less effective.

Model-based object detection approaches for LiDAR, such as \cite{Spinello_Arras_Triebel_Siegwart_2010},  PointNet \cite{PointNet2016}, PointNet++ \cite{PointNetPP2017}, VoxelNet \cite{VoxelNet2018}, and PointPillars \cite{lang2019pointpillarsfastencodersobject} are used to detect larger objects like cars or humans, thus making them insufficient for the detection and tracking of drones.

Unlike model-based object detection, model-free methods detect dynamic objects by comparing consecutive scans, making them more versatile. For instance, \cite{DewanCaselitzTipaldi2016} tracks moving objects by identifying motion patterns between scans, while \cite{Moosmann2013ICRA} employs a Kalman filter combined with Iterative Closest Point (ICP) alignment to track objects in dense 3D LiDAR data. Additionally, \cite{RazlawQuenzelBehnke2019} introduces a method that clusters point clouds using a region growth algorithm and filters objects based on geometric features such as height range and maximum diagonal width. These methods utilize classical rotating LiDARs and also track larger objects like cars or humans.

In \cite{AdaptiveScan2021} a method is proposed for finding MAVs in a point cloud generated by a rosette scanning LiDAR on a moving robot. This approach is further improved by adding a Kalman Filter based tracking method and extensive evaluation in \cite{catalano_2023}. In these papers, the integration time of the LiDAR is dynamically adjusted to adapt to the speed and distance of the target's movement enabling the tracking of MAVs indoors up to about $30\;m$. However, this method does not allow for regaining tracking once lost. Conversely, our aim is to align the rosette scanning pattern LiDAR to ensure continuous centering of the target drone within the sensor FoV, thereby maximizing the number of returned points. Additionally, our approach incorporates a robust tracking method capable of regaining tracking.

In the case of a fixed LiDAR configuration, the usability range in \cite{Hammer_2018} is less than $10\;m$ for the VLP16 sensor and $35-50\; m$ for the HDL-64 sensor. In \cite{Dogru_2022}, it was demonstrated that few measurements are sufficient to track the targets and the maximum detection range was significantly increased with their method. Their approach achieved a maximum detection distance of $70\;m$ with sparse point clouds, and by using retroreflectors, this was extended to $100\;m$. The applied retroreflectors did not increase the size of the drone but increased the number of reflected points from the drone. Our method greatly outperforms this baseline solution in similar weather conditions with a maximum tracking distance of more than $130\;m$ without retroreflector.

LiDAR sensor performance in outdoor environments is highly influenced by weather conditions \cite{Rasshofer}, \cite{Heinzler_2019}. According to \cite{Bijelic_2018} the detection range of LiDAR sensors is limited to $25\;m$  in foggy conditions when the meteorological visibility is under $40\;m$. Based on this a significant reduction in the detection distance and the number of detected points can be expected for outdoor tests under foggy weather conditions. While our experiments confirmed this trend, our method could still track the drone to a little more than $50\;m$. These findings reflect incidental observations rather than a systematic study of weather effects.

%%%%%%%%%%%%%%%%%%%%%%%%%%%%%%%%%%%%%%%%%%%%%%%%%%%%%%%%%%%%%%%%%%%%%%%%%%%%%%%%%%%%%
%                              THE PROPOSED ALGORITHM
%%%%%%%%%%%%%%%%%%%%%%%%%%%%%%%%%%%%%%%%%%%%%%%%%%%%%%%%%%%%%%%%%%%%%%%%%%%%%%%%%%%%%

\section{THE PROPOSED ALGORITHM} \label{proposed_algorithm}

\begin{figure}[t]
\centering
\includegraphics[width=7cm]{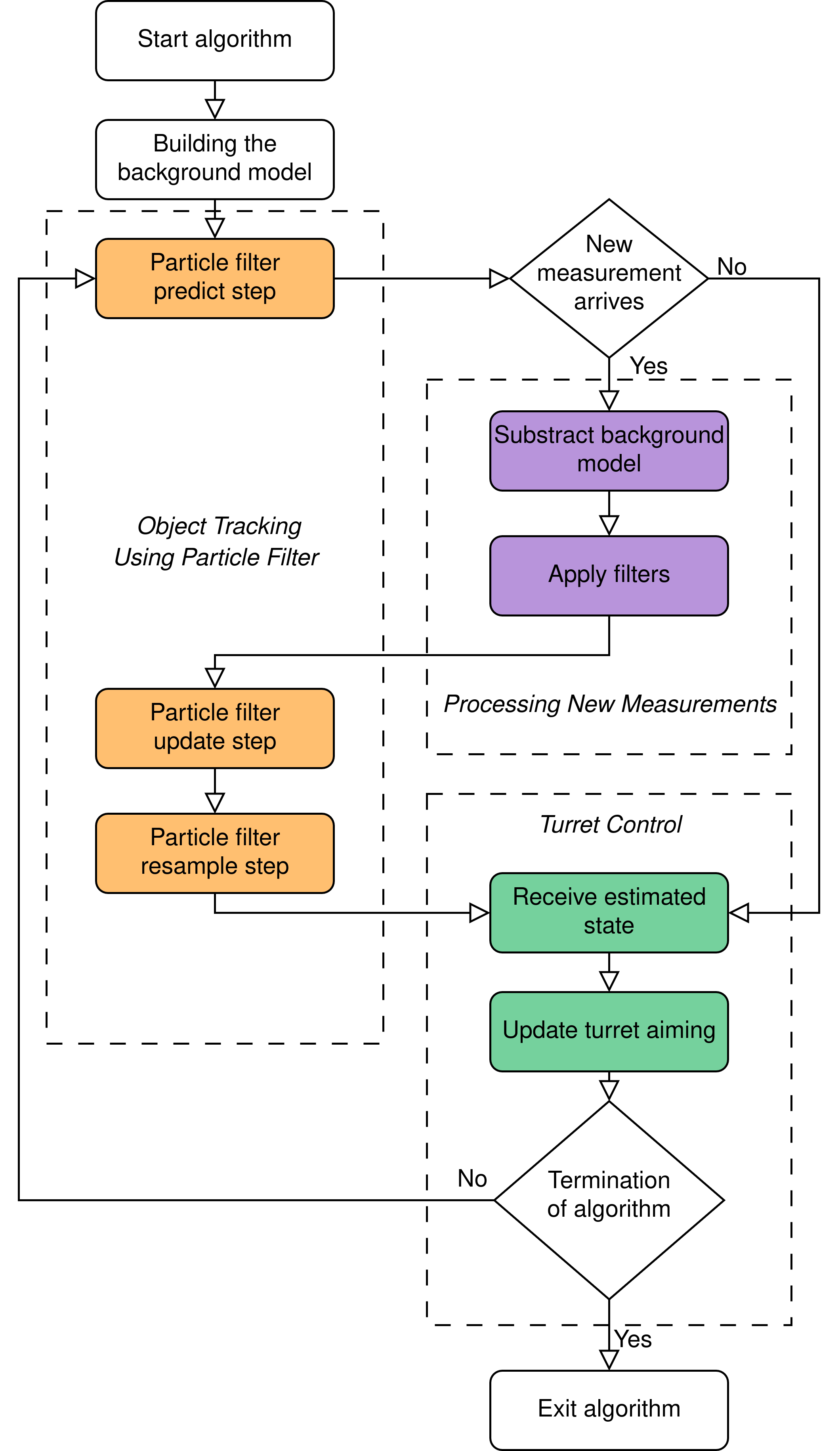}
\caption{Overview of the proposed method.}
\label{fig:algorithm_flow}
\end{figure}

The flowchart of our method is presented in FIGURE \ref{fig:algorithm_flow}, where the main steps of the algorithm are shown. Note on the figure, when the system is installed at a new location, the first step is to build a background model (see Section \ref{background}). This is used to filter the static background from the point cloud to detect the target object.

Next, as part of the object tracking, the particle filter\cite{ParticleFilter_1993} performs the prediction step. If there is a new incoming measurement point cloud, the background model is subtracted to reduce static noise (details are given in Section \ref{processing}). The three steps of the particle filter (predict, update, and resample) are explained in detail in Section \ref{object_tracking_particle_filter}.

Once detected, the drone is kept in the center of the LiDAR's FoV by orienting the pan-tilt turret towards the position computed by the particle filter (see Section \ref{turret_control}). As a result, the measured point density reflected from the drone is maximized.

\subsection{Building the Background Model}
\label{background}

In our application, the background is considered static throughout the drone tracking, hence it is built once at the beginning of the algorithm. Consequently, dynamic objects such as cars, moving foliage, and pedestrians can lead to false positives. To address this, we apply simple mitigation strategies that have proven effective in practice. While we also explored a transformer-based background segmentation approach in \cite{balla2024}, it is not used in this work and falls outside the scope of this paper. The turret is set to scanning mode and performs movements to scan a pre-defined area. Because of the LiDAR's non-repetitive way of scanning, the point density of the whole scene will increase the longer the scanning time is.

The background model is represented with OctoMap\cite{hornung13auro}, which is an efficient probabilistic 3D mapping framework using octrees. The octree tree structure results in fast access times when querying individual voxels.

Based on the position and orientation of the LiDAR, the incoming point cloud is transformed into the static coordinate system of the background model. The points are filtered based on an operator set bounding box, which is set to remove very far away points, the ground, and ground adjacent points, such as pedestrians and cars. After filtering, the point cloud is inserted into the OctoMap of the background model. As the last step of building the background model, after inserting all the collected point clouds, inflation of the background model is performed to account for measurement noise (e.g. vibrations of the turret base) as well as small changes in background objects (such as moving foliage). 

The primary objective of the background removal step is to eliminate the majority of points from the LiDAR measurements that do not correspond to the target. Although some false positives may persist, the tracking algorithm has sufficient robustness to handle these residual points.

\subsection{Processing New Measurements}
\label{processing}

The point measurements are received in the Cartesian frame of the LiDAR and transformed into the World frame based on the turret's pan and tilt angles. The incoming point cloud is filtered during the background building: the ground plane, near points, and far away points are removed based on the same bounding box used at background model building. And finally, the background model is subtracted. There are two additional filters to remove the reflected points that are not close enough to each other and cannot belong to the target object. The filters are the \textit{RadiusOutlierRemoval} and \textit{StatisticalOutlierRemoval} algorithms from the Point Cloud Library (PCL) \cite{Rusu_ICRA2011_PCL}. 

\subsection{Object Tracking Using Particle Filter}
\label{object_tracking_particle_filter}

The drone's position is estimated using a particle filter, which is a recursive Bayesian state estimator that uses discrete particles to approximate the posterior distribution of the estimated state. It accommodates both linear and non-linear process and measurement models, allowing for flexible, arbitrarily shaped probability distributions. This capability is especially useful for MAVs, which exhibit a wide range of motion patterns with abrupt changes in direction and speed across all three dimensions. Additionally, LiDAR measurements enable straightforward updates of the particle states.

The particle filter operates in three main steps: prediction, update, and resampling. The prediction step uses the current state and transition model to predict the next state. In the update step, incoming filtered point cloud data refines the state estimate via Bayes' theorem. Finally, resampling aligns the particles with the posterior distribution, addressing the issue of particle degeneracy.

The particle filter is implemented following Algorithm 2. in \cite{Elfring2021}. Each particle represents a possible state of the MAV. Three different state representations and transition models were implemented and tested, namely: a delta position based estimator, one using a constant velocity motion model and one using a constant acceleration motion model. The state vector $s$ varies depending on the chosen model. For the delta position and constant velocity models $s=[x,y,z,v_x,v_y,v_z]^T$ while for the constant acceleration model $s=[x,y,z,v_x,v_y,v_z,a_x,a_y,a_z]^T$. The position estimation of the drone is calculated by averaging the particle positions. In the upcoming sections we present the equations governing the predict, update, and resample steps of the particle filter for each of the three models.

\subsubsection{Predict}
\label{subsec:pfpredict}

First, the position of each particle is predicted based on its previous position, velocity, acceleration, and added process noise as in (\ref{eq:pospred}). The position prediction noise $p_{\{x,y,z\},k}$ is sampled from a $0$ mean Gaussian distribution with $\sigma_{pos}$ standard deviation, $T$ denotes the period of the particle filter, and $a_{\{x,y,z\},k-1}=0$ in case of the delta position and constant velocity models.

\begin{equation}
\label{eq:pospred}
\begin{gathered}
p_{\{x,y,z\},k}\sim \mathcal{N}(0,\sigma_{pos}^2)\\
\begin{bmatrix}
x_k \\
y_k \\
z_k 
\end{bmatrix}=
\begin{bmatrix}
x_{k-1} + v_{x,k-1} \cdot T + 0.5 \cdot a_{x,k-1} \cdot T^2 + p_{x,k}\\
y_{k-1} + v_{y,k-1} \cdot T + 0.5 \cdot a_{y,k-1} \cdot T^2 + p_{y,k}\\
z_{k-1} + v_{z,k-1} \cdot T + 0.5 \cdot a_{z,k-1} \cdot T^2 + p_{z,k}
\end{bmatrix}
\end{gathered}
\end{equation}

The velocities are computed in two different ways. In case of the delta position model, the term is dependent on the status of the tracking. The tracking can be considered stable when the standard deviation of the particle positions ($\sigma_{particles}$) is around the parameter $\sigma_{pos}$. The $\sigma_{particles}$ value increases with $\sigma_{pos}$ plus the additional effect of the velocity in every iteration when the drone is lost.

The MAV is considered to be successfully tracked if $\sigma_{particles}$ remains below a threshold defined as $\sigma_{threshold} = 1.5 \cdot \max(\sigma_{pos}, \sigma_m)$, where $\sigma_m$ is the measurement uncertainty. This condition ensures that when the measurement is less reliable than the position estimate, the resulting estimate reflects the greater uncertainty. We scale the velocity with the scaling factor $\rho$ that is inversely proportional to that uncertainty.

\begin{equation}
\rho = 
\begin{cases}
1, & \text{if } \sigma_{particles} < \sigma_{threshold} \\
\frac{\sigma_{pos}}{\sigma_{particles}}, & \text{otherwise}
\end{cases}
\end{equation}
\begin{equation}
\begin{bmatrix}
v_{x,k} \\
v_{y,k} \\
v_{z,k} 
\end{bmatrix}=
\begin{bmatrix}
\rho \cdot (x_k-x_{k-1})/T \\
\rho \cdot (y_k-y_{k-1})/T \\
\rho \cdot (z_k-z_{k-1})/T
\end{bmatrix}
\end{equation}

In case of the constant velocity and constant acceleration models, the velocity is computed as in (\ref{eq:constvel1}-\ref{eq:constvel2}), where  $a_{\{x,y,z\},k-1}=0$ in case of the constant velocity model.

\begin{equation}
\label{eq:constvel1}
p_{\{v_x,v_y,v_z\},k}\sim \mathcal{N}(0,\sigma_{vel}^2)
\end{equation}
\begin{equation}
\label{eq:constvel2}
\begin{bmatrix}
v_{x,k} \\
v_{y,k} \\
v_{z,k} 
\end{bmatrix}=
\begin{bmatrix}
v_{x,k-1} + a_{x,k-1} \cdot T + p_{v_x,k}\\
v_{y,k-1} + a_{y,k-1} \cdot T + p_{v_y,k}\\
v_{z,k-1} + a_{z,k-1} \cdot T + p_{v_z,k}
\end{bmatrix}
\end{equation}

Finally, in case of the constant accelartion model the acceleration is updated as in (\ref{eq:constacc1}-\ref{eq:constacc2}).

\begin{equation}
\label{eq:constacc1}
p_{\{a_x,a_y,a_z\},k}\sim \mathcal{N}(0,\sigma_{acc}^2)
\end{equation}
\begin{equation}
\label{eq:constacc2}
\begin{bmatrix}
a_{x,k} \\
a_{y,k} \\
a_{z,k} 
\end{bmatrix}=
\begin{bmatrix}
a_{x,k-1} + p_{a_x,k}\\
a_{y,k-1} + p_{a_y,k}\\
a_{z,k-1} + p_{a_z,k}
\end{bmatrix}
\end{equation}

\subsubsection{Update}
The weight of each particle is updated when a new measurement arrives based on (\ref{eq:weightupdate}). The new weight is calculated using a Gaussian distribution with $\sigma_m$ standard deviation, which represents the uncertainty of the lidar sensor measurement. This distribution is evaluated at $d_i$, which is the distance between the $i$th particle and the closest LiDAR measurement point. The weights are then normalized (\ref{eq:normalization}) so that the sum of $\tilde{w_i}$ will add up to 1, where $n$ is the number of particles.

\begin{equation}
\label{eq:weightupdate}
w_{i} = \mathcal{N}(d_i ; 0, \sigma_m^2)
\end{equation}
\begin{equation}
\label{eq:normalization}
\tilde{w_i}=\frac{w_i}{\sum_{j}^{n}w_j}
\end{equation}

\subsubsection{Resample}
After every update step, the particles are resampled using their weights as the probability of being drawn for the next iteration.

\subsubsection{Particle initialization and reset}
The particles are initialized randomly in a cylinder, whose parameters (3D location, radius, and height) are set based on the detection area. The particles can be reset to a new cylinder during runtime, thus giving the operator the capability to reset the tracking to another object. During tracking $\sigma_{particles}<\sigma_{threshold}$, which means 2 objects are only mistaken if they are closer than $\sigma_{threshold}$, which is almost certainly a collision of the objects.

\subsection{Turret control}
\label{turret_control}

Turret control is done in two ways: in initialization or tracking mode. During initialization mode, the pan-tilt turret collects data points for a specified time in a predefined area to build the background model. In tracking mode, the control commands are calculated based on the current position of the turret and the estimated position of the drone. The tracking mode aims to keep the drone in the middle of the FoV to maximize the density of reflected points from the drone. In case there are multiple drones in the FoV, the algorithm follows the one closest to the location of the particles, which  can be reset by the operator.

The pan ($\alpha$) and tilt ($\beta$) goal angles of the turret are calculated based on the current drone position estimate as in (\ref{eq:angles}).
\begin{equation}
    \label{eq:angles}
    \begin{aligned}
        \alpha &= \tan^{-1}\left(\frac{y_{mav}}{x_{mav}}\right) \\
        \beta &= -\tan^{-1}\left(\frac{z_{mav}}{\sqrt{(x_{mav} )^2 + (y_{mav} )^2}}\right)
    \end{aligned}
\end{equation}

\subsection{Implementation details}

The Robot Operating System (ROS) was used to integrate and manage communication between program components, offering a reliable framework for modular nodes. Four nodes were implemented: a LiDAR wrapper node to fix point cloud conversion issues; a pan-tilt turret wrapper node to translate MAV position estimates into turret commands and send them to the turret's position controller; a background builder node for the Octomap background model; and a MAV detection and tracking node that holds the core algorithm, including the particle filter and other Extended Kalman Filter (EKF) implementations for benchmarking. Note that the source code is made available\footref{footnote:git}.

%%%%%%%%%%%%%%%%%%%%%%%%%%%%%%%%%%%%%%%%%%%%%%%%%%%%%%%%%%%%%%%%%%%%%%%%%%%%%%%%%%%%%
%                               EXPERIMENTS
%%%%%%%%%%%%%%%%%%%%%%%%%%%%%%%%%%%%%%%%%%%%%%%%%%%%%%%%%%%%%%%%%%%%%%%%%%%%%%%%%%%%%

\section{Experiments}\label{experiments}

In the following subsections, we explain the parameter settings and discuss the experimental results, including a comparison with other methods. We first present and validate the equations for setting the parameters of the different motion models of the particle filter in Subsection \ref{subsec:paramoptim}. In Subsection \ref{subsec:ablation}, we quantify the effect of the different parts of our algorithm. Then, in Subsection \ref{subsec:tracking_comp}, an extensive comparison is carried out on four recordings, comparing our algorithm with other SOTA methods \cite{Dogru_2022,catalano_2023}. A real-world field test is presented in Subsection \ref{subsec:realtest}. The accuracy of the proposed method is investigated in \ref{subsec:accuracy}. Finally, we show the track regain capability of our algorithm in \ref{subsec:lostandfound}.

In all of the experiments, the turret used was the WidowX Dual XM430 Pan \& Tilt turret by Trossen Robotics, and the integration time of the LIVOX AVIA sensor was set to $100\;ms$.

\subsection{Motion model parameter selection}\label{subsec:paramoptim}

The particle filter relies on several key parameters: the number of particles $n$ and noise parameters $\sigma_{pos}$, $\sigma_{vel}$, $\sigma_{acc}$, $\sigma_{m}$, which govern how particles evolve and interact with the measurements. We justify our parameter choices based on computational constraints, drone dynamics, and experimental validation.

To ensure real-time operation, we evaluated the effect of particle count $n$ on computational load using the Tracy Profiler\footnote{\url{https://github.com/wolfpld/tracy}}. The filter was benchmarked on a single thread of an Intel Core i7 (8th generation) desktop CPU. The LiDAR callback can take up a worst-case execution time of $52\;ms$ (but usually takes much less: $1.45\;ms$ mean with $1.57\;ms$ standard deviation), regardless of the number of particles so each filter step (prediction, update, and resampling combined) had to complete within $48ms$ to maintain a $10\;Hz$ filter rate that is needed to process the LiDAR data at $10\;Hz$. TABLE~\ref{tab:timings} presents the  mean, standard deviation, and worst-case execution times of the particle filter across different particle counts and the worst-case full execution time with the LiDAR callback, which has to be under $100\;ms$ to be able to run at $10\;Hz$.

\begin{table}[!ht]
\caption{\textbf{Execution times with different particle numbers. Worst-case values are significant outliers, as indicated by the mean and standard deviation.}}
\label{tab:timings}
\renewcommand{\arraystretch}{1.3}
\setlength{\tabcolsep}{3pt} 
\centering
\begin{tabular}{lccccccc}
\textbf{n $[-]$} & \textbf{10000} & \textbf{9000} & \textbf{8000} & \textbf{7000} & \textbf{6000} & \textbf{5000} & \textbf{4000} \\ \hline
Mean PF $[ms]$ & 20.60 & 17.30 & 16.03 & 15.38 & 12.5 & 10.85 & 10.55 \\
Std PF $[ms]$ & 2.52 & 3.56 & 1.88 & 4.54 & 3.19 & 3.13 & 4.81 \\
\makecell[l]{Worst-case \\ PF $[ms]$} & 103.00 & 88.32 & 81.38 & 59.11 & 50.67 & 34.62 & 33.16 \\
\makecell[l]{Worst-case \\ full $[ms]$} & 155.00 & 140.32 & 133.38 & 111.11 & 102.67 & 86.62 & 85.16 
\end{tabular}
\end{table}

The particles approximate the probability distribution of the model. The accuracy of this approximation improves with increasing particle count, asymptotically approaching the true distribution. The number was maximized while still enabling real-time operations. Based on these results, we set $n = 5000$ to ensure real-time performance while maintaining a sufficiently accurate representation of the posterior distribution.

For context, we report the EKF execution times of the two SOTA methods. The mean, standard deviation, and worst-case execution times are $(0.12\;[ms];\,0.11\;[ms];\,1.73\;[ms])$ for \cite{Dogru_2022} and $(0.12\;[ms];\,0.16\;[ms];\,3.03\;[ms])$ for \cite{catalano_2023}. While these results indicate noticeably lower execution times, the difference is not significant in the context of the overall pipeline and is justified by the performance improvements outlined in the following Subsections.

We tested all three motion models for particle prediction. Each model predicts particles based on different assumptions and therefore requires different parameter sets. 
Initial tests revealed that while the delta position and constant velocity models performed comparably, the constant acceleration model consistently underperformed even with small $\sigma_{acc}$ values. This was due to over-inflation of the particle cloud and poor handling of abrupt directional changes, issues caused by the assumption of continuous acceleration. As a result, it was not considered for further evaluation.

Based on the results of these initial experiments, we derived a unified set of equations for initializing the parameters of all models, grounded in the expected dynamics of the drone. This unification promotes consistency and simplifies the tuning process.

The equations below define $\sigma_{pos}$, $\sigma_{vel}$, $\sigma_{acc}$, and $\sigma_m$ in terms of known or measurable drone dynamics, namely expected maximum speed $v_{max}$ and expected maximum acceleration $a_{max}$.

\begin{equation}
\label{eq:pfparams}
\begin{aligned}
    \sigma_{pos} &= \frac{v_{max} \cdot T}{3} \\
    \sigma_{vel} &= \frac{a_{max} \cdot T}{3} \\
    \sigma_{acc} &= 0 \\
    \sigma_{m} &= \frac{s_{diag}}{2}
\end{aligned}
\end{equation}

In (\ref{eq:pfparams}), $T$ is the filter's time step and $s_{diag}$ is the diagonal distance between the drone’s motors. The choice of dividing by 3 ensures that approximately 99.74\% of the particles remain within the physically plausible range (assuming a Gaussian distribution). The $\sigma_{acc}$ parameter is set to $0$ because there is no constantly changing acceleration in the drone's dynamics. Finally, $\sigma_m$ was chosen similarly to \cite{catalano_2023}.

\begin{figure*}[h]
\centering
\includegraphics[width=\linewidth]{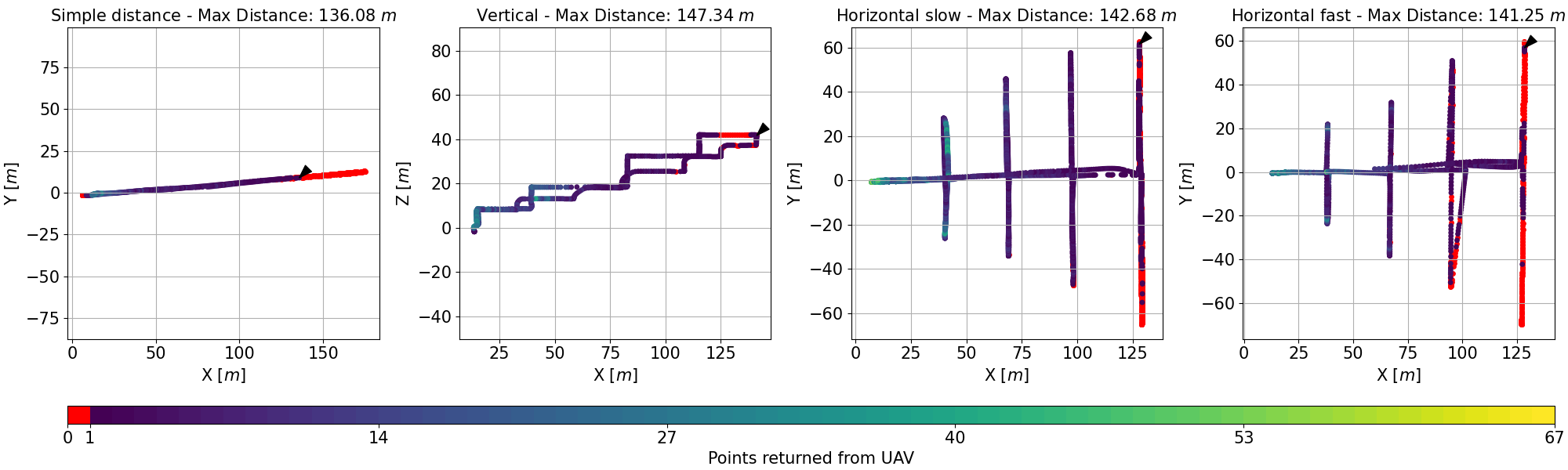}
\caption{Returned point counts and trajectories are shown for the four different evaluation flights. The color indicates the number of MAV points in the point cloud from the LIVOX sensor, with red representing zero returns. The 'Simple distance', 'Horizontal slow', and 'Horizontal fast' trajectories were flown at constant altitude and are therefore plotted from a top-down perspective. The 'Vertical' trajectory was flown in a line varying the altitude, and is plotted from a side view. All trajectories were designed such that, at the maximum distance, no points are returned for extended periods. The maximum detection distance is indicated above each figure, with a small black arrow also marking this point along the trajectory.}
\label{fig:trajectories}
\end{figure*}

To validate these parameters, we conducted flight experiments with a DJI Phantom 4 ($s_{diag}=0.4\;m$). The drone was flown along a challenging trajectory, including high-speed passes perpendicular to the LiDAR’s axis and out-of-range segments to test edge cases that can be seen in FIGURE \ref{fig:trajectories} on the far right. Ground Truth (GT) positions were obtained via GPS, streamed in real-time to the processing laptop using UGCS\footnote{\url{https://www.ugcs.com/}}, where the turret was moved based on these GT measurements. Despite minor latency, the setup provided reliable ground truth for evaluation.

The maximum observed velocity and acceleration were approximately $15\;m/s$ and $15\;m/s^2$, leading to the following parameters based on (\ref{eq:pfparams}): $\sigma_{pos}=0.5\;m$, $\sigma_{vel}=0.5\;m/s$, $\sigma_{acc}=0\;m/s^2$, and $\sigma_{m}=0.2\;m$. This speed and acceleration represent the upper limit at which our pilot was still comfortable controlling the drone, and is therefore considered a good approximation of realistic conditions in practical scenarios.

We then performed a grid search to evaluate the system’s sensitivity to parameter variations:

\begin{equation}
    \begin{aligned}
        \sigma_{pos} &= [0.4,0.5,0.6]\;[m] \\
        \sigma_{vel} &= [0.0,0.4,0.5,0.6]\;[m/s] \\
        \sigma_{acc} &= [0.0,0.05,0.1,0.15,0.2]\;[m/s^2] \\
        \sigma_{m} &= [0.1,0.2,0.3]\;[m]
    \end{aligned}
\end{equation}

Each parameter combination was evaluated across 5 trials to account for the non-deterministic nature of the particle filter. We used two metrics: (1) \textbf{maximum tracking distance}, defined as the average distance of the furthest correctly identified MAV point cloud (within $4\;m$ of GPS ground truth), and (2) \textbf{track coverage}, the percentage of the flight where the MAV was correctly tracked.

\begin{figure}[t]
\centering
\includegraphics[width=0.85\linewidth]{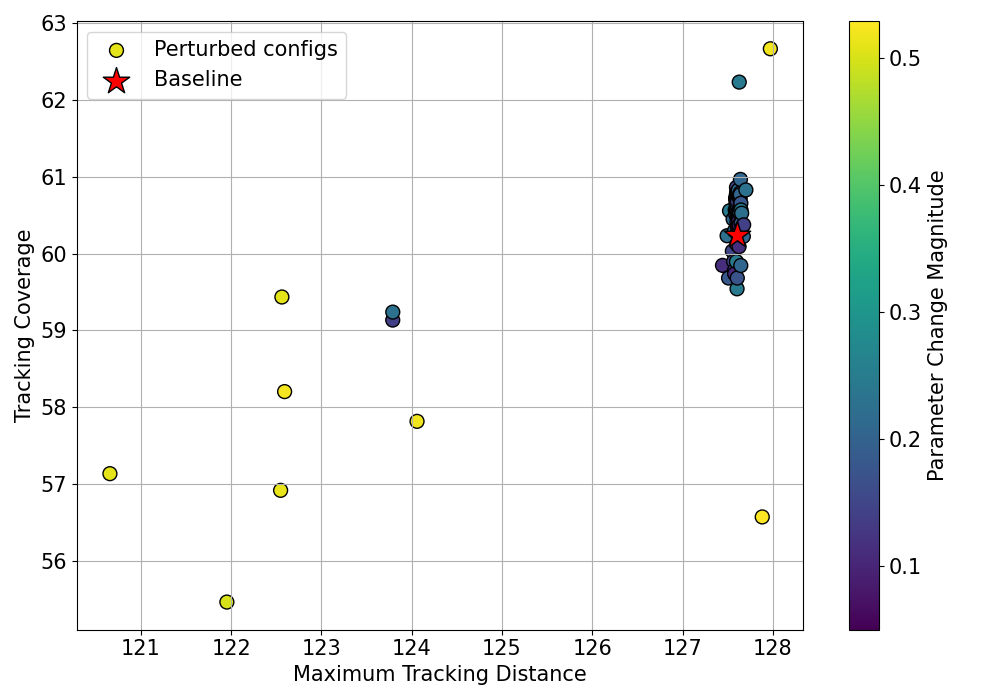}
\caption{Performance of different parameter combinations in terms of tracking distance and coverage. Colors represent distance from the analytically derived parameter set. }
\label{fig:paramgrid}
\end{figure}

FIGURE \ref{fig:paramgrid} shows a subset of the results for the constant acceleration model for illustration purposes. Points are color-coded by the Euclidean distance from the analytical parameter set given by equations (\ref{eq:pfparams}). Results indicate that the filter is robust to small deviations in parameter values, and performance drops only under large perturbations.

The narrow performance variance across trials and the existence of a stable parameter zone around the analytical parameter set indicate that the algorithm is stable under parameter perturbations.

All code, datasets, and evaluation scripts are publicly available at our git repository. This includes raw LiDAR point clouds, GPS ground truth, parameter configuration files, and plotting tools.

\subsection{Ablation study}
\label{subsec:ablation}

\begin{figure}[t]
\centering
\includegraphics[width=0.95\linewidth]{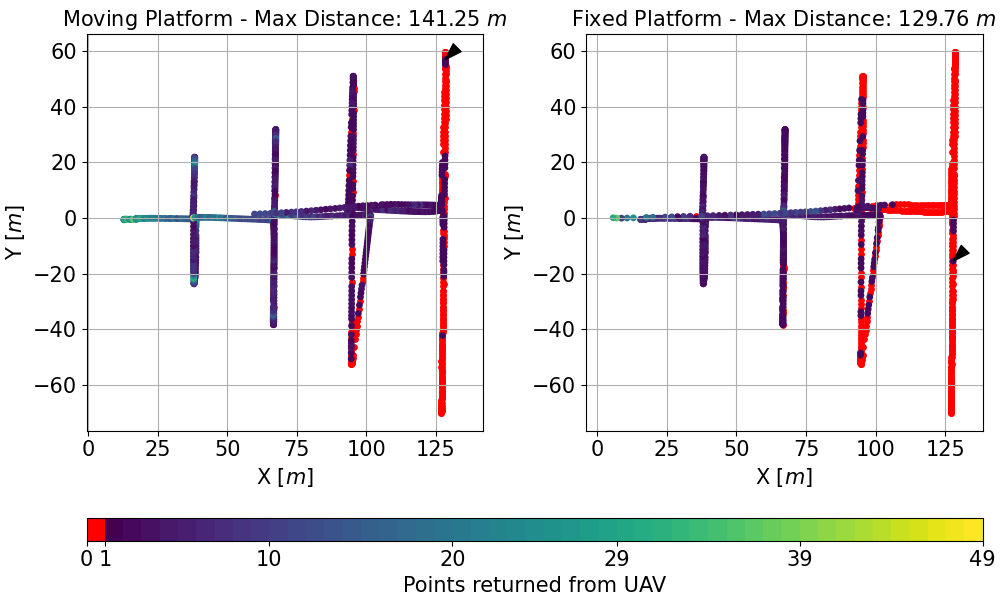}
\caption{Returned point counts and trajectory for the moving and fixed platforms for the 'Horizontal fast' trajectory (the left of this figure is the same as the far right of FIGURE \ref{fig:trajectories}). A notable difference is observed in the furthest sections of the trajectory, where the stationary LiDAR returns almost no points beyond $100\;m$.}
\label{fig:ablation_study}
\end{figure}

In this subsection, we evaluate the individual components of our proposed algorithms to assess their contributions to overall performance. To isolate the effect of the moving LiDAR, we performed a control experiment where a static, non-moving LIVOX AVIA sensor was positioned next to the moving platform and recorded the same trajectory used in Subsection \ref{subsec:paramoptim}. The number of returned points for both configurations is shown in FIGURE \ref{fig:ablation_study}, where a significant difference can be observed, particularly in the furthest sections of the trajectory.

We compare the proposed delta position and constant velocity motion model PFs with the SOTA method of Dogru et al. \cite{Dogru_2022} using an EKF with a constant velocity motion model. TABLE \ref{tab:ablation_study} presents the results of our ablation experiments. Each experiment was repeated 5 times to account for the non-deterministic nature of the methods.

The baseline maximum tracking distance of $70\;m$ is the reported result of \cite{Dogru_2022}, which was recorded using a circular rotating Velodyne VLP-16 LiDAR that tracked a drone with $s_{diag}=0.45\;m$, slightly larger than our DJI Phantom 4. When using the same EKF algorithm within the proposed tracking system, which utilizes the rosette scanning pattern of the LIVOX AVIA sensor, the tracking distance improves by approximately $15\;\%$, which we attribute to the sensor's superior performance. Our particle filters extend this distance by an additional $30\;\%$ in the fixed LiDAR configuration.

In case of the moving platform, the maximum tracking distances are increased by another $30\;\%$, and are comparable across the EKF and both PFs. However, tracking coverage with the PFs is more than five times higher. We attribute the additional $30\;\%$ performance gain in maximum tracking distance to our strategy of moving the platform to maintain the drone within the center of the FoV. Overall, the best combination of maximum tracking distance and track coverage is achieved by the delta position PF.

\begin{table}[t]
\caption{\textbf{Results of the ablation study.}}
\label{tab:ablation_study}
\renewcommand{\arraystretch}{1.3}
\setlength{\tabcolsep}{3pt} 
\centering
\begin{tabular}{lcc}
\makecell[l]{\textbf{Method}} & \makecell[c]{\textbf{Coverage $[\%]$} \\ mean / std} & \makecell[c]{\textbf{Max Dist. $[m]$} \\ mean / std} \\ \hline
\makecell[l]{Fix, EKF \cite{Dogru_2022}\\ reported with VLP-16} & $N.A.$ & $\sim70 / N.A.$ \\
\makecell[l]{Fix, EKF \cite{Dogru_2022}} & $6.60/0.03$ & $80.3/0.0$ \\
\makecell[l]{Fix, PF constant velocity} & $43.91 / 1.37$ & $103.79 / 0.88$ \\
\makecell[l]{Fix, PF delta position} & $51.59 / 4.47$ & $102.83 / 0.70$ \\
\makecell[l]{Moving, EKF \cite{Dogru_2022}} & $12.70 / 0.42$ & $131.70 / 5.33$ \\
\makecell[l]{Moving, PF constant velocity} & $68.42 / 1.00$ & $129.16 / 0.38$  \\
\makecell[l]{Moving, PF delta position} & $71.35 / 6.07$ & $129.68 / 0.36$  
\end{tabular}
\end{table}

\subsection{Tracking performance comparision}\label{subsec:tracking_comp}

The previously tuned particle filters, utilizing constant velocity and delta position motion models, were benchmarked against the EKF tracking methods presented by Dogru et al. \cite{Dogru_2022} and Catalano et al. \cite{catalano_2023}. These comparisons were conducted within the proposed rosette scanning pattern LiDAR system, where the turret was controlled using GT GPS positions streamed from the drone, ensuring the drone remained centered within the FoV. Four different trajectories were considered and recorded on a sunny day. These are shown in FIGURE \ref{fig:trajectories}, along with the corresponding returned point counts.

TABLE \ref{tab:cov_and_max} summarizes the results of this set of experiments. Each test was executed 5 times to account for the non-deterministic nature of the filters. We report both the mean and standard deviation for maximum tracking distance and track coverage metrics.

\begin{table*}[t]
\caption{\textbf{Maximum detection distance and track coverage results on four outside trajectories. Bold font denotes the best result in its column. Our PF using the delta position model performs best in terms of coverage in all scenarios, while it matches the SOTA in maximum tracking distance.}}
\label{tab:cov_and_max}
\renewcommand{\arraystretch}{1.4}
\setlength{\tabcolsep}{3pt}
\centering
\begin{tabular}{l|cc|cc|cc|cc}
& \multicolumn{2}{c|}{\textbf{Simple distance}} & \multicolumn{2}{c|}{\textbf{Vertical}} & \multicolumn{2}{c|}{\textbf{Horizontal slow}} & \multicolumn{2}{c}{\textbf{Horizontal fast}} \\ 
\textbf{Method} & \multicolumn{1}{c|}{\makecell[c]{\textbf{Coverage $[\%]$} \\ mean / std}} & \makecell[c]{\textbf{Max Dist. $[m]$} \\ mean / std} 
& \multicolumn{1}{c|}{\makecell[c]{\textbf{Coverage $[\%]$} \\ mean / std}} & \makecell[c]{\textbf{Max Dist. $[m]$} \\ mean / std} 
& \multicolumn{1}{c|}{\makecell[c]{\textbf{Coverage $[\%]$} \\ mean / std}} & \makecell[c]{\textbf{Max Dist. $[m]$} \\ mean / std} 
& \multicolumn{1}{c|}{\makecell[c]{\textbf{Coverage $[\%]$} \\ mean / std}} & \makecell[c]{\textbf{Max Dist. $[m]$} \\ mean / std} \\ \cline{1-9}
\makecell[l]{EKF \cite{catalano_2023}} & 2.60 / 0.06 & 14.09 / 0.00 & 4.41 / 0.21 & 16.65 / 0.09 & 2.61 / 0.03 & 10.88 / 0.32 & 0.00 / 0.00 & 0.00 / 0.00 \\
\makecell[l]{EKF \cite{catalano_2023} mod.} & 9.83 / 0.21 & 32.51 / 0.40 & 20.38 / 0.11 & 59.42 / 0.55 & 31.09 / 0.01 & 71.44 / 0.00 & 15.20 / 0.00 & 44.75 / 0.00 \\
\makecell[l]{EKF \cite{Dogru_2022}} & 37.80 / 1.56 & 131.23 / 0.00 & 7.55 / 0.03 & 51.21 / 0.00 &  14.08 / 0.03 & \textbf{138.18 / 0.00} & 12.70 / 0.42 & \textbf{131.71 / 5.33} \\
\makecell[l]{PF constant\\velocity} & 43.02 / 0.84 & 131.23 / 0.00 & 55.84 / 0.51 & \textbf{140.18 / 9.15} & 73.05 / 0.63 & 132.02 / 2.28 & 68.42 / 1.00 & 129.16 / 0.38\\
PF delta position & \textbf{55.19 / 7.53} & \textbf{131.97 / 2.02} & \textbf{80.35 / 5.25} & 129.93 / 0.28 & \textbf{74.72 / 3.20} & 135.57 / 1.61 & \textbf{71.35 / 6.07} & 129.68 / 0.36
\end{tabular}
\end{table*}

The EKF in \cite{catalano_2023} was originally tuned for indoor scenarios with static LiDAR placement, where the drones operate at lower speeds. Their method employs a Nearest-Neighbor Search (NNS) algorithm for finding the next possible drone measurement with a fixed radius ($r=s_{diag}/2 = 0.2\;m$) to associate measurements. We benchmarked our algorithm against their original method (EKF \cite{catalano_2023} in TABLE \ref{tab:cov_and_max}) and a modified one, where we increased the NNS search radius (EKF \cite{catalano_2023} mod. in TABLE \ref{tab:cov_and_max}). It was increased to $r=1.5\;m$, since this is the expected maximum travel distance based on the drone dynamics in one iteration of the filter. The EKF filters were all re-initialized at a specific time, when the LiDAR measurement only contained points from the drone. This adjustment was made because their original papers used different approaches for the initial detection, so to ensure a fair comparison, a proper initialization was provided. If the drone was successfully tracked for at least five consecutive frames, results are included in the table, otherwise, a value of 0.00 is reported.

The results indicate that our PF-based algorithms achieve consistently better track coverage than the EKF-based methods. The EKF introduced in \cite{catalano_2023} with the increased NNS radius performs well at close ranges. However, once the drone exits the expected radius, the tracker fails and cannot recover. \cite{Dogru_2022} partially solves this problem by dynamically setting the NNS radius, which allows their EKF to occasionally reacquire the drone. Nevertheless, after that, their EKF prediction is usually inaccurate, and the drone is lost again in a few frames. This is reflected in the lower coverage values shown in TABLE \ref{tab:cov_and_max}, except for the 'simple distance' trajectory, where the motion is constant and predictable, enabling successful recovery. Our PF-based solutions adapt more effectively to the combination of sensor modality, drone dynamics, and missing measurements, thus having significantly better coverage across all scenarios.

Due to the unpredictable nature of drone motion, out of the two particle filters, the delta position-based solution performs consistently better in terms of coverage. It is somewhat interesting that it also outperforms the model-based one on the 'simple distance' trajectory, where the speed is almost constant. This is likely because the high velocity noise (which is needed to track the drone when it changes direction quickly) results in bigger position errors, ultimately reducing its coverage.

Overall, we found that our particle filter with delta position motion model matches the SOTA EKF-based solution in terms of maximum tracking distance, while providing significantly better coverage. This is particularly important in real-world scenarios, where the turret is moved based on the detection of the filter to keep the drone in the middle of the FoV.

\subsection{Field test}\label{subsec:realtest}

We performed two measurements with the presented delta position PF method under two different weather conditions (on a sunny and a foggy day). The purpose of the validation is to test the track movement and to determine the detection distances and number of detected points in real-world conditions with turret input from the filter.

The same DJI Phantom 4 was used for these experiments. The measurements were carried out in the factory state of the drone, we did not improve the detection range by using special paint or other methods. For a video of the outdoor experiments, kindly check the video attachment here: \url{https://youtu.be/ghfjqAnDuag}.

\begin{figure}[t]
\centering
\includegraphics[width=0.95\linewidth]{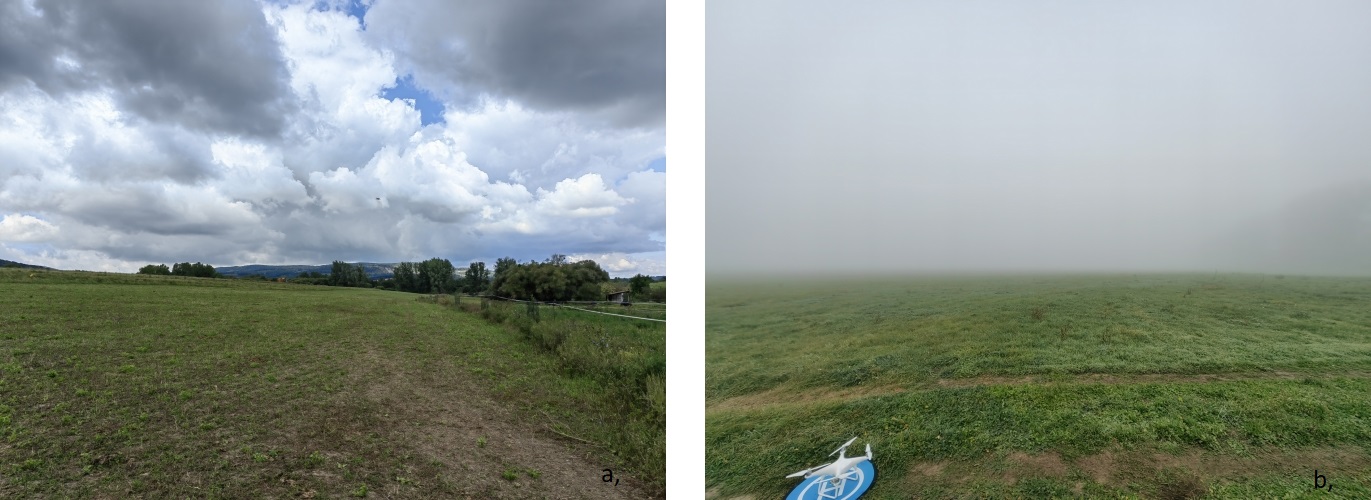}
\caption{Weather on the test days. Sunny experiment with clouds on the left and the foggy experiment with severely limited viewing conditions on the right.}
\label{fig:weather_test_day}
\end{figure}

A sunny day was chosen for the first and a foggy day for the second measurement day, when visibility was also severely limited (see FIGURE \ref{fig:weather_test_day}).

During both tests, the drone was moving under such conditions that both horizontal and vertical distances could be analyzed. In both cases, the controlled pan-tilt turret followed the drone correctly, thus the drone tracking was successful.

\begin{figure}[t]
\centering
\begin{subfigure}{0.95\linewidth}
    \includegraphics[width=8cm]{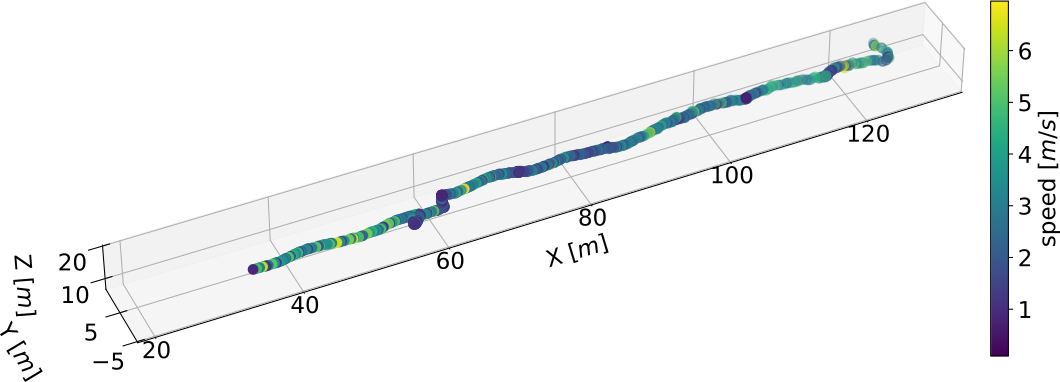}
    \caption{Sunny}
    \label{subfig:outside_sunny}
\end{subfigure}
\begin{subfigure}{0.95\linewidth}
    \includegraphics[width=8cm]{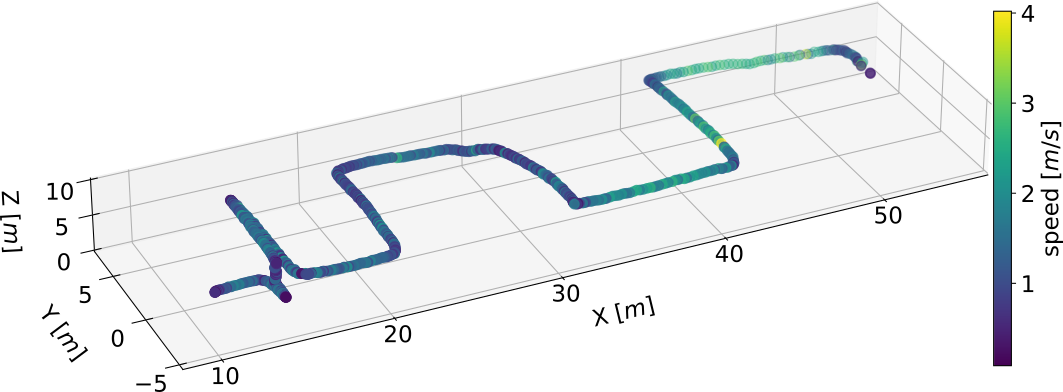}
    \caption{Foggy}
    \label{subfig:outside_foggy}
\end{subfigure}
\caption{Trajectory and speed of the drone on a sunny and foggy day. The turret was at the origin of the coordinate system.}
\label{fig:outside}
\end{figure}

The drone's trajectory was successfully tracked up to a distance of $128\;m$ in the sunny test environment (FIGURE \ref{subfig:outside_sunny}), and up to a distance of $58\;m$ in the foggy test environment (FIGURE \ref{subfig:outside_foggy}). As explained earlier in subsection \ref{subsec:pfpredict}, we consider the drone tracked when $\sigma_{particles}<0.3\;m$. It is important to note that we had some occasional detections at distances of up to $145\;m$ on the sunny day.

In both scenarios, we calculated maximum speeds of a little more than $6\;m/s$ from the tracking positions and timestamps. 

\begin{figure}[t]
\centering
\includegraphics[width=8cm]{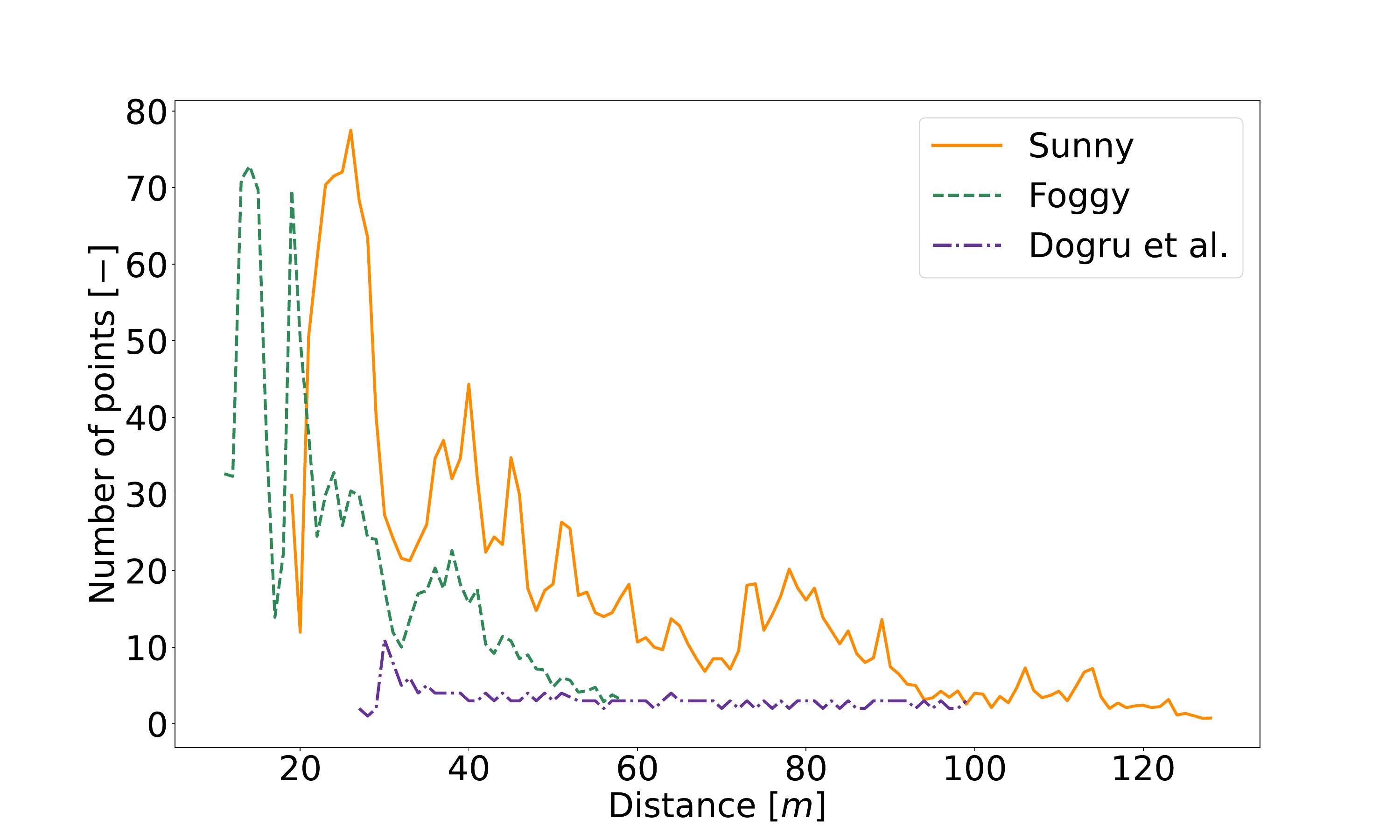}
\caption{Number of detected points depending on distance. The proposed algorithm returns considerably more points in every distance range than Dogru et al.\cite{Dogru_2022}.}
\label{fig:compare_detected_point}
\end{figure}

Another important aspect is the number of detected points, which greatly impacts the robustness of tracking. In sunny weather, except for the last part of the detection range, the number of detected points in the result is $3$ to $5$ times higher than in the current SOTA (see FIGURE \ref{fig:compare_detected_point}). This can be attributed partly to the sensor choice and partly to our strategy of maintaining the drone within the center of the FoV based on our findings in Subsection \ref{subsec:ablation}.

In the case of foggy weather, the detection distance was significantly reduced, as expected in advance based on the literature\cite{Bijelic_2018}. While a detailed investigation of the performance in different weather conditions is beyond the scope of this work, we include this observation as an incidental but noteworthy outcome of our study.

\subsection{Accuracy}\label{subsec:accuracy}

These experiments were conducted within the MIMO arena \cite{MIMO_new_2022} within the HUN-REN SZTAKI Institute for Computer Science and Control incorporating an OptiTrack motion capture system in a \(7 \times 8 \; m^{2}\) office space with a height of \(3.5 \; m\), which can provide sub-millimeter accurate, 240 Hz tracking data for bodies with IR retroreflective markers. For the target of the experiment a Crazyflie 2.0 MAV by Bitcraze was used, which measures $s_{diag}=12\;cm$. The expected maximum velocity and acceleration were $1.2\;m/s$ and $1.2\;m/s^2$ respectively, so the following noise parameters were set for the PF: $\sigma_{pos}=0.04\;m$, $\sigma_{vel}=0.04\;m$, $\sigma_{m}=0.06\;m$.

\begin{figure}[t]
\centering
\includegraphics[width=8cm]{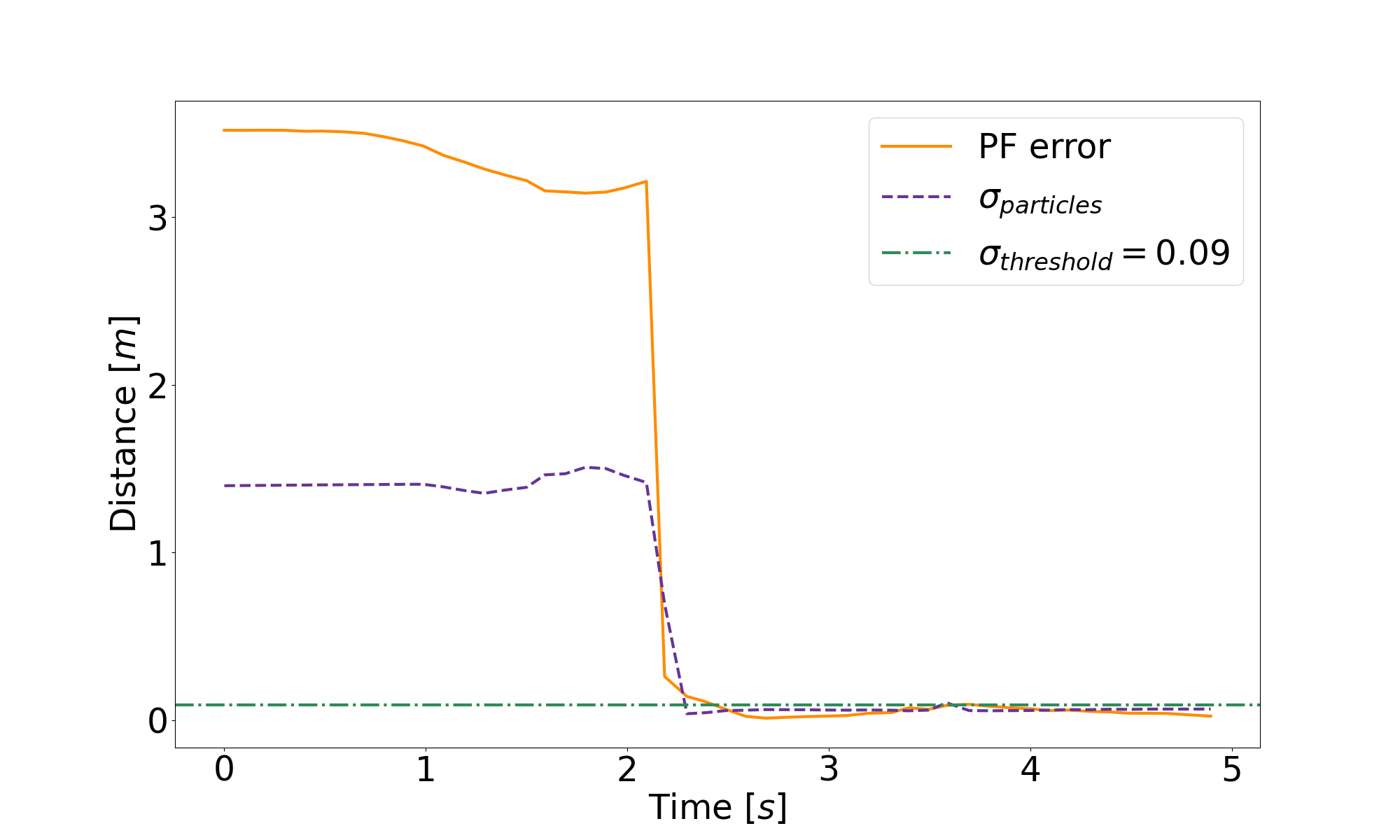}
\caption{Initial detection error and $\sigma_{particles}$ of the particle filter. The particle filter locks onto the drone almost instantaneously after receiving the first measurement at around 2 seconds. $\sigma_{threshold}=\sigma_m\cdot1.5=0.09\;m$ is also plotted.}
\label{fig:initial_detection_error}
\end{figure}

FIGURE \ref{fig:initial_detection_error} shows the moment the drone gets tracked by the delta position PF, by plotting the error as the distance of the estimated state to the true state in meters of a representative detection. The average of the standard deviation of the particles in each dimension is also plotted (denoted as $\sigma_{particles}$).

While the drone is out of the FoV of the LiDAR sensor, the particles remain idle, only influenced by the added noise in the predict step which results in a slowly inflating particle cloud with a slowly rising $\sigma_{particles}$. In this case, the MAV is resting on the floor and, because of its small size, is hidden inside the background model. As the drone starts hovering and measurements of it arrive at around $2\;s$ the particles gather and allow for the first good estimation. Parallel with this, the $\sigma_{particles}$ of the estimation drops sharply. This means the drone is tracked.

\begin{figure}[t]
\centering
\includegraphics[width=7.5cm]{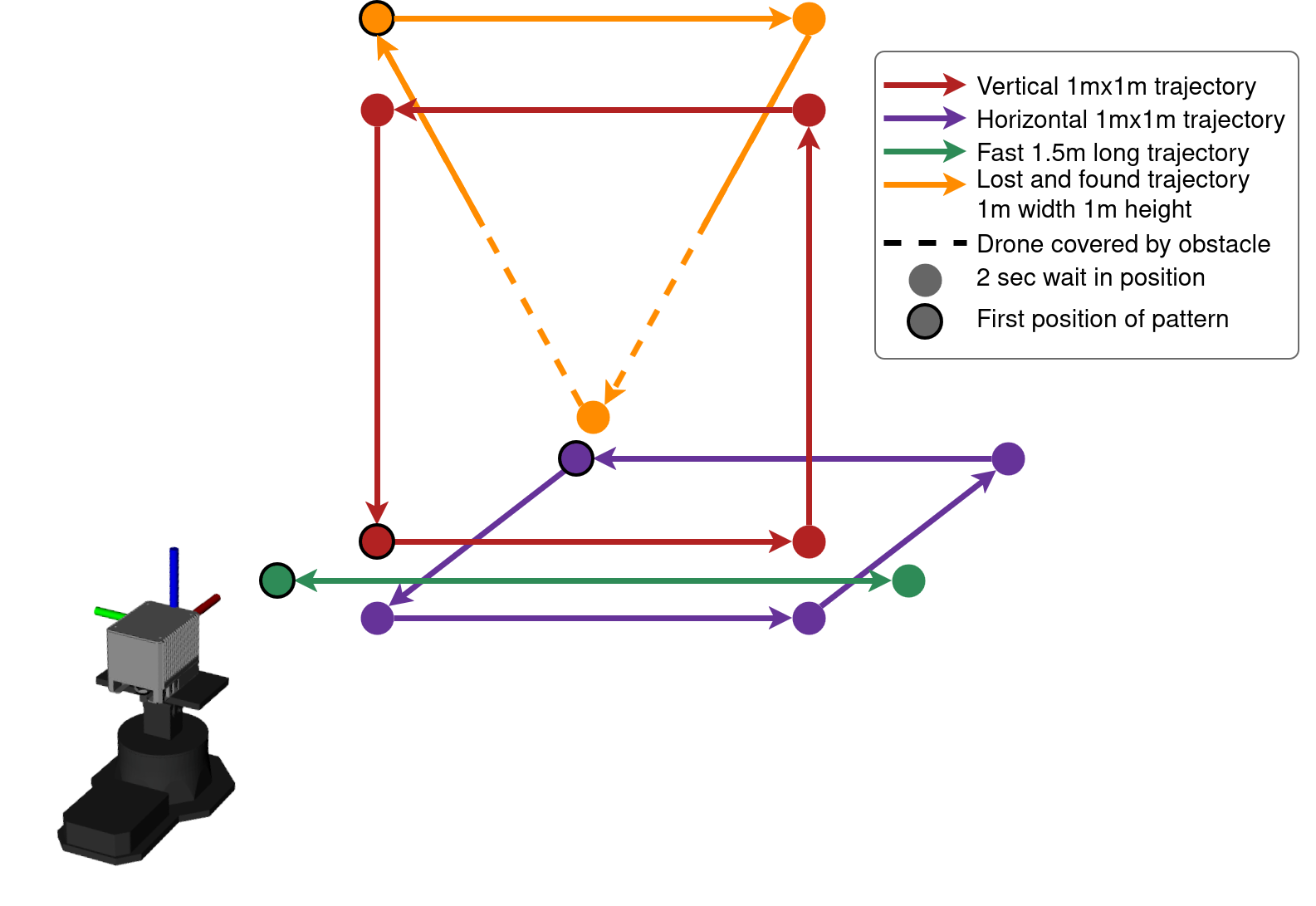}
\caption{Different drone testing flight patterns with different colors (Vertical, Horizontal, Fast, and 'Lost and found'). The outlined circles show the starting position of each trajectory while the arrows denote the moving direction of the drone. The dashed part of the arrow means that the MAV view is obstructed from the LiDAR's point of view by an obstacle. The size of each trajectory is in the legend of the figure.}
\label{fig:test_flight_patterns}
\end{figure}

For calculating the accuracy, the MAV was flown following certain flight patterns (described in FIGURE \ref{fig:test_flight_patterns}) in a repeated manner. The ground truth data was obtained by the motion capture system within the MIMO arena by recording the pose of the LIVOX sensor and the position of the drone and aligning the LIVOX measurements with the MIMO arena using the recorded pose. The patterns in FIGURE \ref{fig:test_flight_patterns} were drawn from the point of view of the LiDAR sensor where the middle of each trajectory in the horizontal direction was in front of the sensor with a minimum distance of around $3\;m$. Each circle represents a stationary wait time of $2\;s$ and the one with a black outline represents the first position of the respective pattern. The trajectory represented by the arrows in-between was flown in $3\;s$ in case of the vertical, horizontal, and 'lost and found' pattern which resulted in a maximum speed of about $0.6\;m/s$, while it was flown in $2.25\;s$ in case of the fast pattern which resulted in a maximum speed of around $1.2\;m/s$. In each experiment, the drone flies to the first point and then it flies the pattern 3 times in the case of the vertical and horizontal patterns and 4 times in the case of the others.

\begin{table*}[ht]
\caption{\textbf{Accuracy comparison for different tracking methods. RMSE and 95 \% confidence intervals are reported across five runs for each trajectory. Our delta position PF consistently achieves the lowest errors, which is particularly apparent in the ‘Fast’ and ‘Lost and found’ scenarios.}}
\label{tab:accuracy}
\renewcommand{\arraystretch}{1.4}
\setlength{\tabcolsep}{3pt}
\centering
\begin{tabular}{l|cc|cc|cc|cc}
& \multicolumn{2}{c|}{\textbf{Horizontal}} & \multicolumn{2}{c|}{\textbf{Vertical}} & \multicolumn{2}{c|}{\textbf{Fast}} & \multicolumn{2}{c}{\textbf{Lost and found}} \\ 
\textbf{Method} & \multicolumn{1}{c|}{\textbf{RMSE $[m]$}} & \makecell[c]{\textbf{CI 95\% $[m]$} \\ lower / upper}
& \multicolumn{1}{c|}{\textbf{RMSE $[m]$}} & \makecell[c]{\textbf{CI 95\% $[m]$} \\ lower / upper} 
& \multicolumn{1}{c|}{\textbf{RMSE $[m]$}} & \makecell[c]{\textbf{CI 95\% $[m]$} \\ lower / upper} 
& \multicolumn{1}{c|}{\textbf{RMSE $[m]$}} & \makecell[c]{\textbf{CI 95\% $[m]$} \\ lower / upper} \\ \cline{1-9}
\makecell[l]{EKF \cite{catalano_2023}} & 0.078 & 0.076 / 0.081 & 0.595 & 0.577 / 0.614 & 2.604 & 2.474 / 2.749 & 1.880 & 1.810 / 1.955 \\
\makecell[l]{EKF \cite{Dogru_2022}} & 0.072 & 0.070 / 0.075 & 0.071 & 0.069 / 0.073 & 0.239 & 0.227 / 0.252 & 0.800 & 0.771 / 0.833 \\
\makecell[l]{PF constant\\velocity} & 0.080 & 0.078 / 0.082 & 0.077 & 0.075 / 0.080 & 0.156 & 0.147 / 0.166 & 0.643 & 0.619 / 0.669 \\
PF delta position & \textbf{0.054} & 0.053 / 0.056 & \textbf{0.057} & 0.055 / 0.059 & \textbf{0.132} & 0.125 / 0.139 & \textbf{0.292} & 0.281 / 0.304 
\end{tabular}
\end{table*}

TABLE \ref{tab:accuracy} presents a comparison of the accuracy results for the different tracking filters. As before, all tests were conducted five times to account for the non-deterministic nature of the filters. To ensure a fair comparison, all filters were properly initialized. We report the Root Mean Square Error (RMSE) over all five runs, calculated using the Euclidean distance as defined in (\ref{eq:rmse}) where $(\hat{x}_i,\hat{y}_i,\hat{z}_i)$ denotes the the filter's prediction, $(x_i,y_i,z_i)$ denotes the GT position, and $n$ is the number of GT measurements. The $95\;\%$ confidence intervals were computed analytically using the Chi-Squared method. The 'Lost and found' trajectory is discussed separately in Subsection \ref{subsec:lostandfound}.

\begin{equation}
    \label{eq:rmse}
    \text{RMSE} = \sqrt{ \frac{1}{n} \sum_{i=1}^{n} \left( \sqrt{(\hat{x}_i - x_i)^2+(\hat{y}_i - y_i)^2+(\hat{z}_i - z_i)^2} \right)^2}
\end{equation}

In these experiments, we also evaluate the proposed method against the EKF implementation from Catalano et al.\cite{catalano_2023} and the EKF approach proposed by Dogru et al.\cite{Dogru_2022}. In the case of \cite{catalano_2023}, we used their original NNS search radius, which was tuned for indoor scenarios and did not modify it for this evaluation.

For the 'Horizontal' trajectory, the observed results are consistent with those reported in the respective papers. However, in the 'Vertical' and 'Fast' trajectories, the EKF in \cite{catalano_2023} performs considerably worse than the other methods. This poor performance is likely due to the motion model they employed: a Constant Turn Rate and Velocity (CTRV) model extended with a constant vertical velocity component equal to that of the planar motion. This results in degraded estimation along the vertical axis. In the 'Fast' trajectory, tracking is lost almost immediately, which accounts for the high RMSE values there.

The proposed delta position PF consistently outperforms all other methods across all evaluated scenarios. This demonstrates that our approach not only matches the SOTA in terms of maximum tracking distance and surpasses it significantly in terms of track coverage, but is also the most accurate overall.

\begin{figure}[ht!]
\centering
\includegraphics[width=8cm]{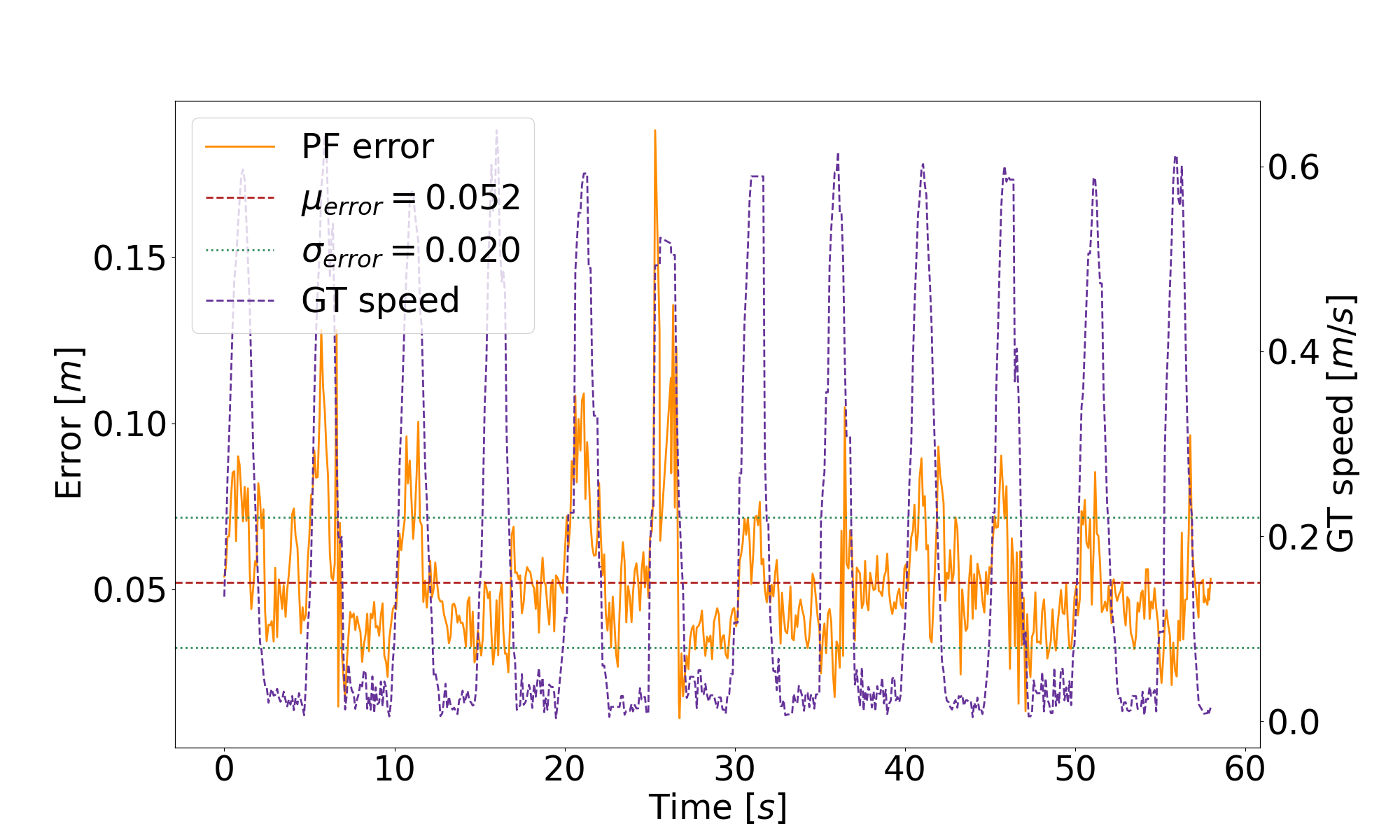}
\includegraphics[width=8cm]{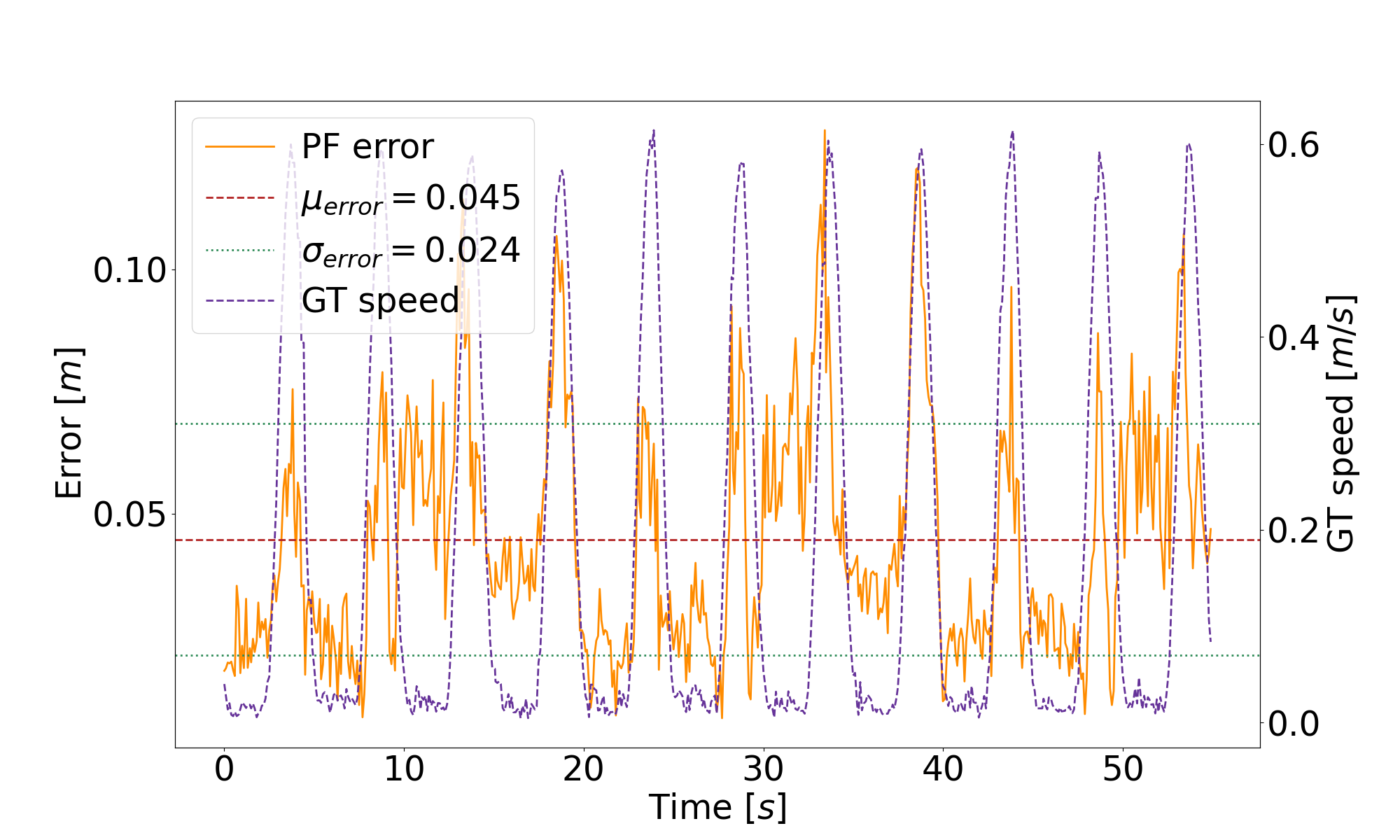}
\includegraphics[width=8cm]{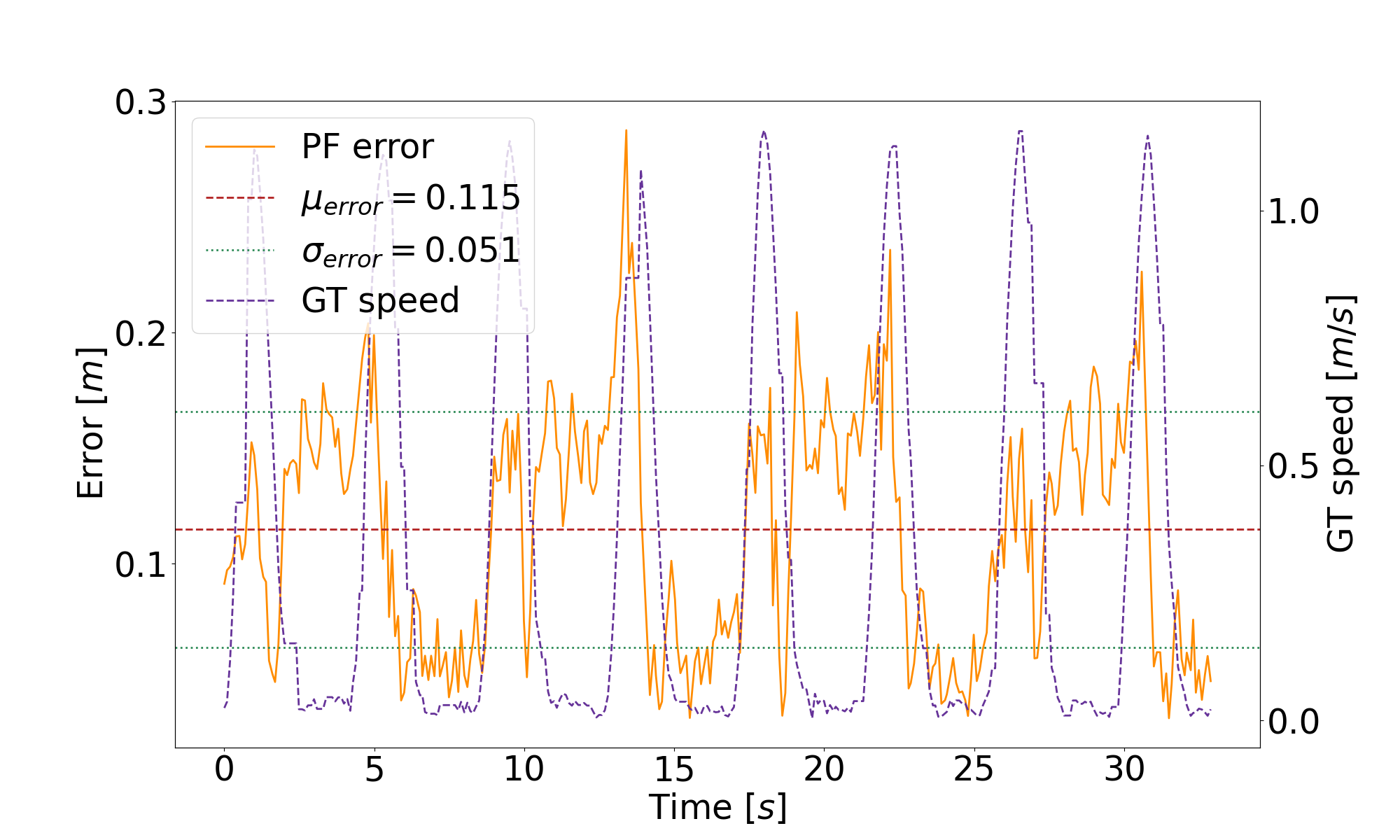}
\caption{Position error of the particle filter plotted together with ground truth speed from the MIMO arena for the Vertical (top), Horizontal (middle), and Fast (bottom) patterns. The mean ($\mu_{error}$) and standard deviation ($\pm \sigma_{error}$) of the particle filter error are also plotted.}
\label{fig:accuracy_test}
\end{figure}

In FIGURE \ref{fig:accuracy_test}, the Euclidean error of the delta position particle filter (first of the 5 runs in every scenario) is plotted with the ground truth speed calculated from the MIMO measurements. It is visible that the tracking is stable over time with an average position error of around $5\;cm$ in the case of the horizontal and vertical measurements, while it is around $11\;cm$ for the fast pattern.

It can be seen on all the plots of FIGURE \ref{fig:accuracy_test}, that the error of the particle filter is dependent on the speed. When the drone is at a waypoint, not moving, the error of the filter is around $3$ to $5\;cm$, while at the peak speeds of $1.2\;m/s$ of the fast pattern (not counting the one outlier), the error is around $20\;cm$. This dynamic difference of about $15\;cm$ in the error can mostly be explained by the different delays in the pipeline. The internal processing time of the LiDAR firmware is $2\;ms$ based on the data-sheet, the LiDAR scan integration time is $100\;ms$, and there is some additional overhead in the formatting, stamping, and publishing of the data to ROS topics. The LiDAR callback took $1.45\;ms$ on avarage while the particle filter (predict, update, and resample steps including the background subtraction) took $10.85\; ms$. The whole delay is about $120\;ms$, in which the drone travels roughly $14.4\;cm$, which accounts for most of the dynamic error.

It is important to note that the size of the Crazyflie drone ($s_{diag} = 12\;cm$) is on the same order of magnitude as the observed error. The LiDAR only returns points from the side of the drone facing the sensor, and the markers used by the OptiTrack system for tracking are intentionally placed asymmetrically. As a result, the true center of the drone does not exactly coincide with the measured GT position. Although these factors are difficult to quantify precisely, they all contribute to the overall error.

Overall, when the drone is stationary, the error remains small and consistent. However, during movement, a dynamic error arises that increases linearly with speed. This affects the tracking accuracy but does not compromise its robustness.

\subsection{Losing and regaining Tracking}\label{subsec:lostandfound}

An advantage of utilizing a particle filter is that the detection and tracking properties are part of the same routine, which leads to good performance in terms of regaining tracking after the target is lost.

\begin{figure}[t]
\centering
\begin{subfigure}{0.45\linewidth}
    \includegraphics[width=3.3cm]{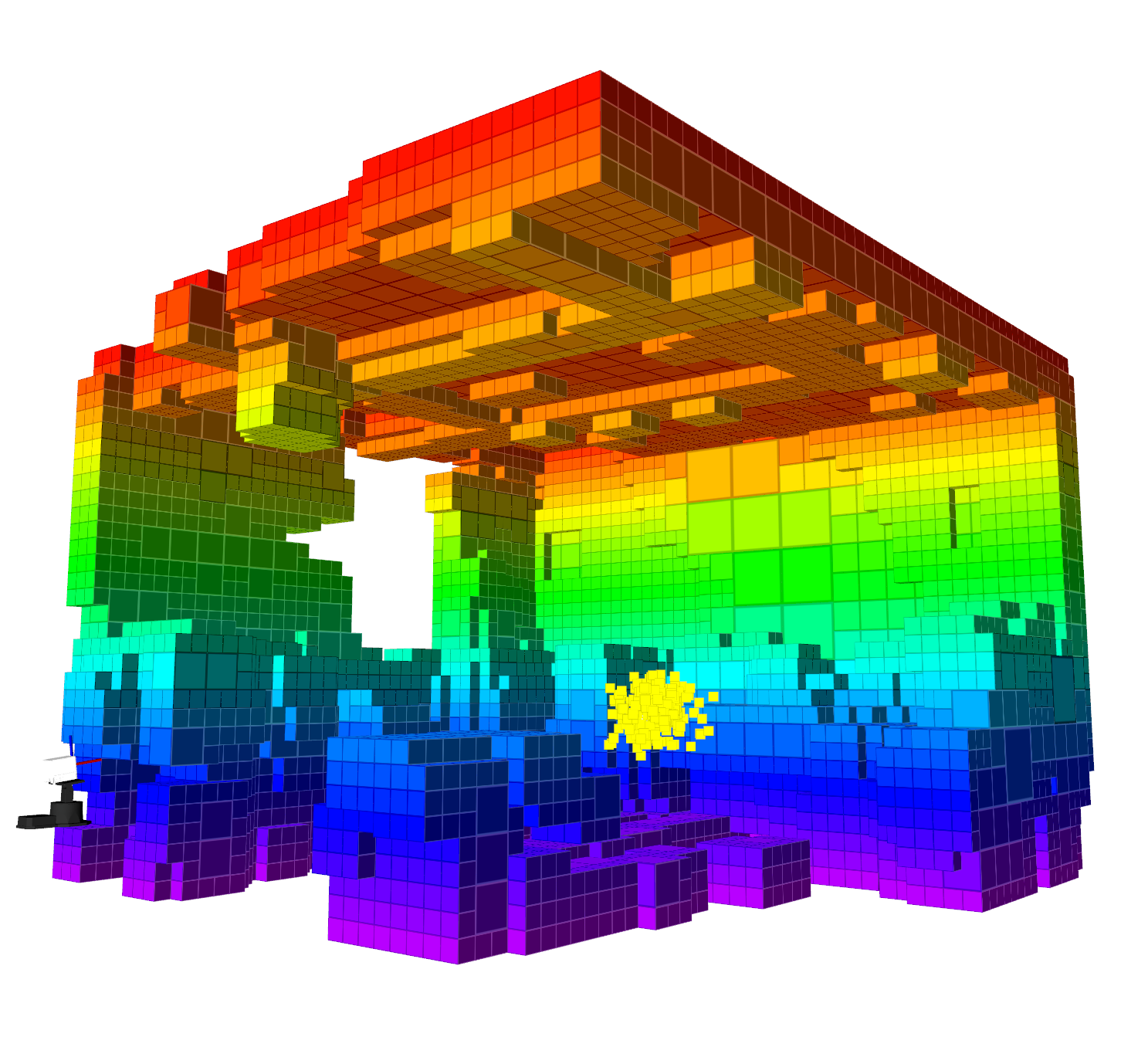}
    \caption{Stable before loss}
    \label{subfig:lost&found1}
\end{subfigure}
\begin{subfigure}{0.45\linewidth}
    \includegraphics[width=3.3cm]{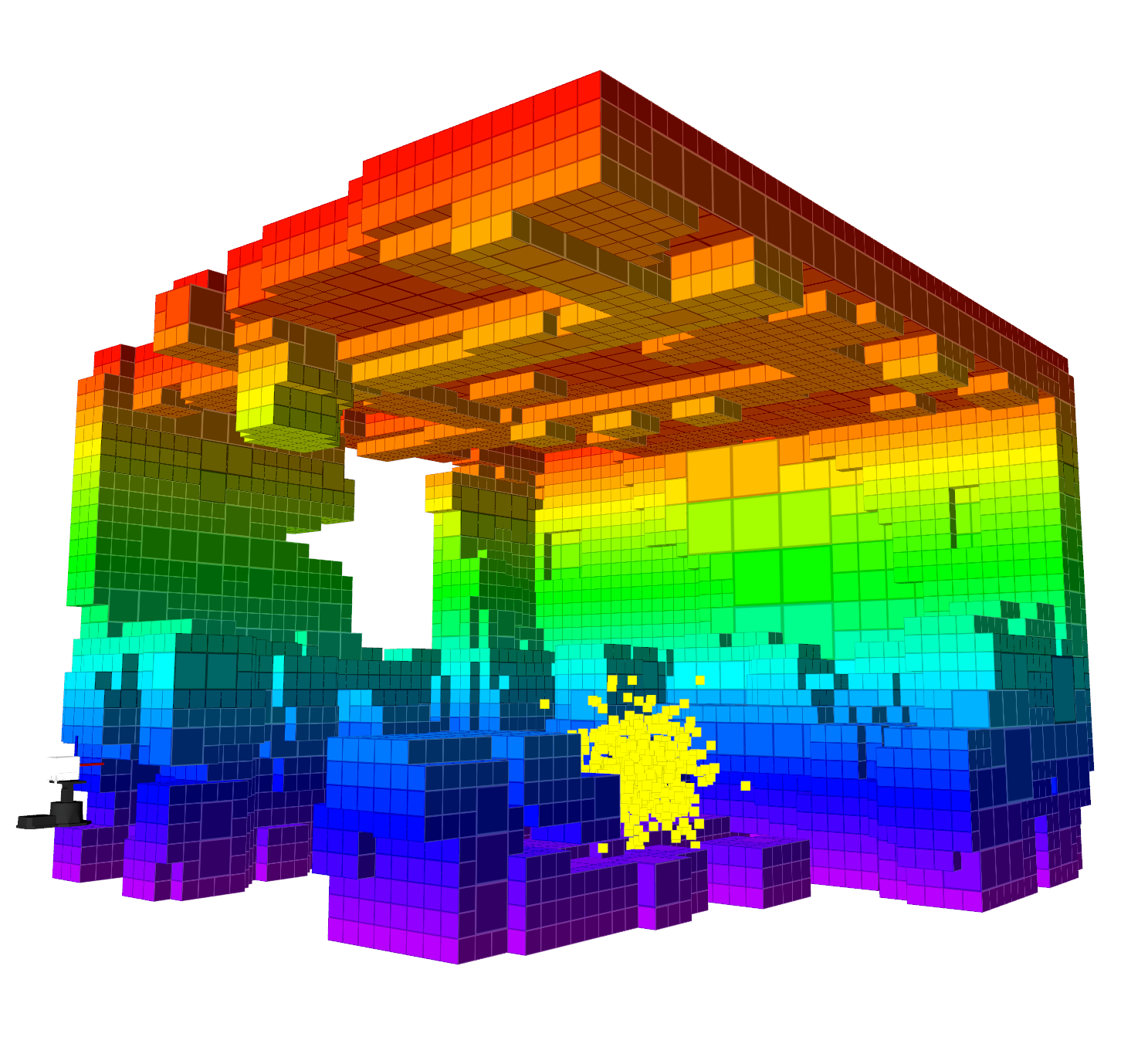}
    \caption{Right after loss}
    \label{subfig:lost&found2}
\end{subfigure}
\begin{subfigure}{0.45\linewidth}
    \includegraphics[width=3.3cm]{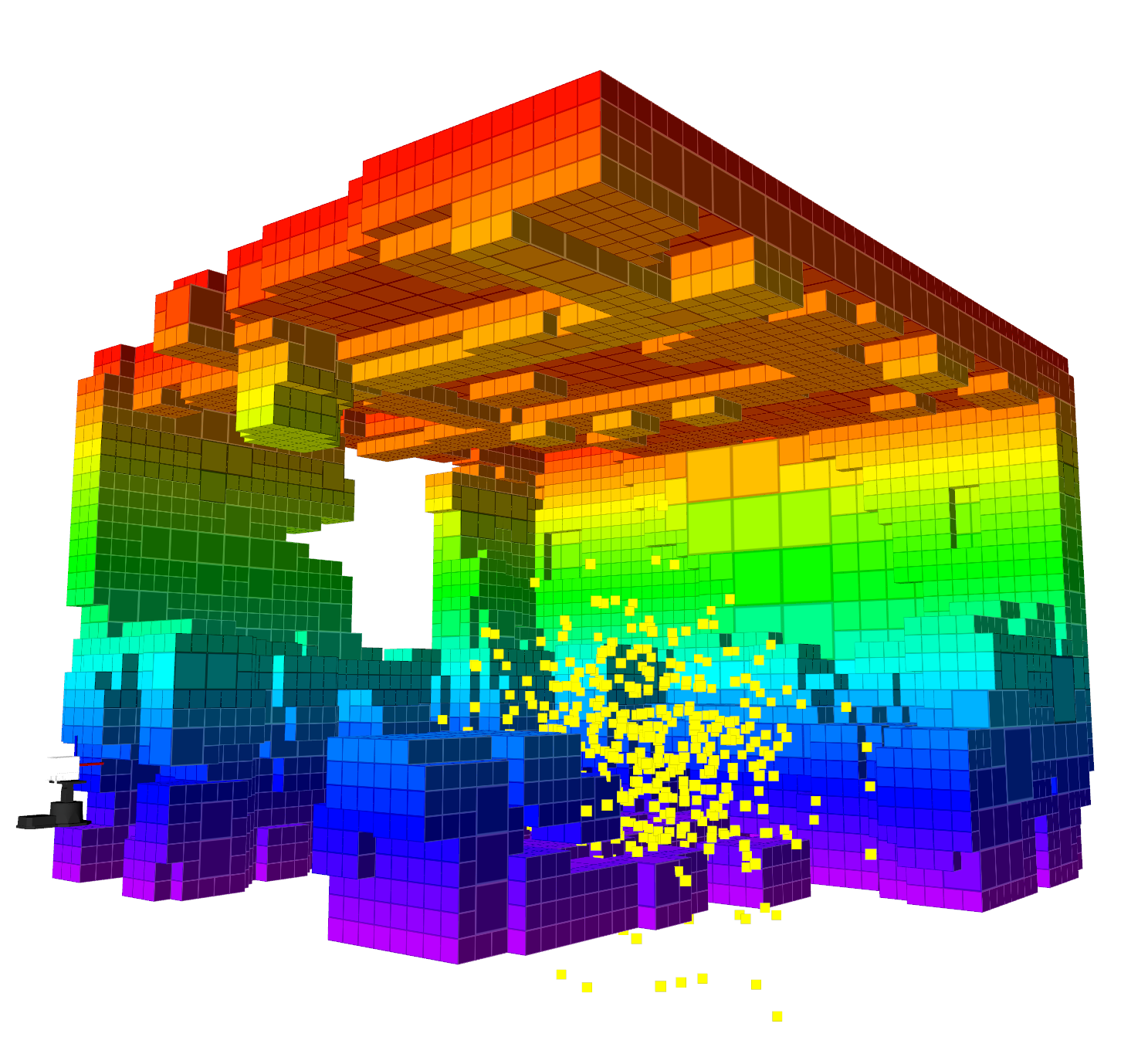}
    \caption{Lost for some time}
    \label{subfig:lost&found3}
\end{subfigure}
\begin{subfigure}{0.45\linewidth}
    \includegraphics[width=3.3cm]{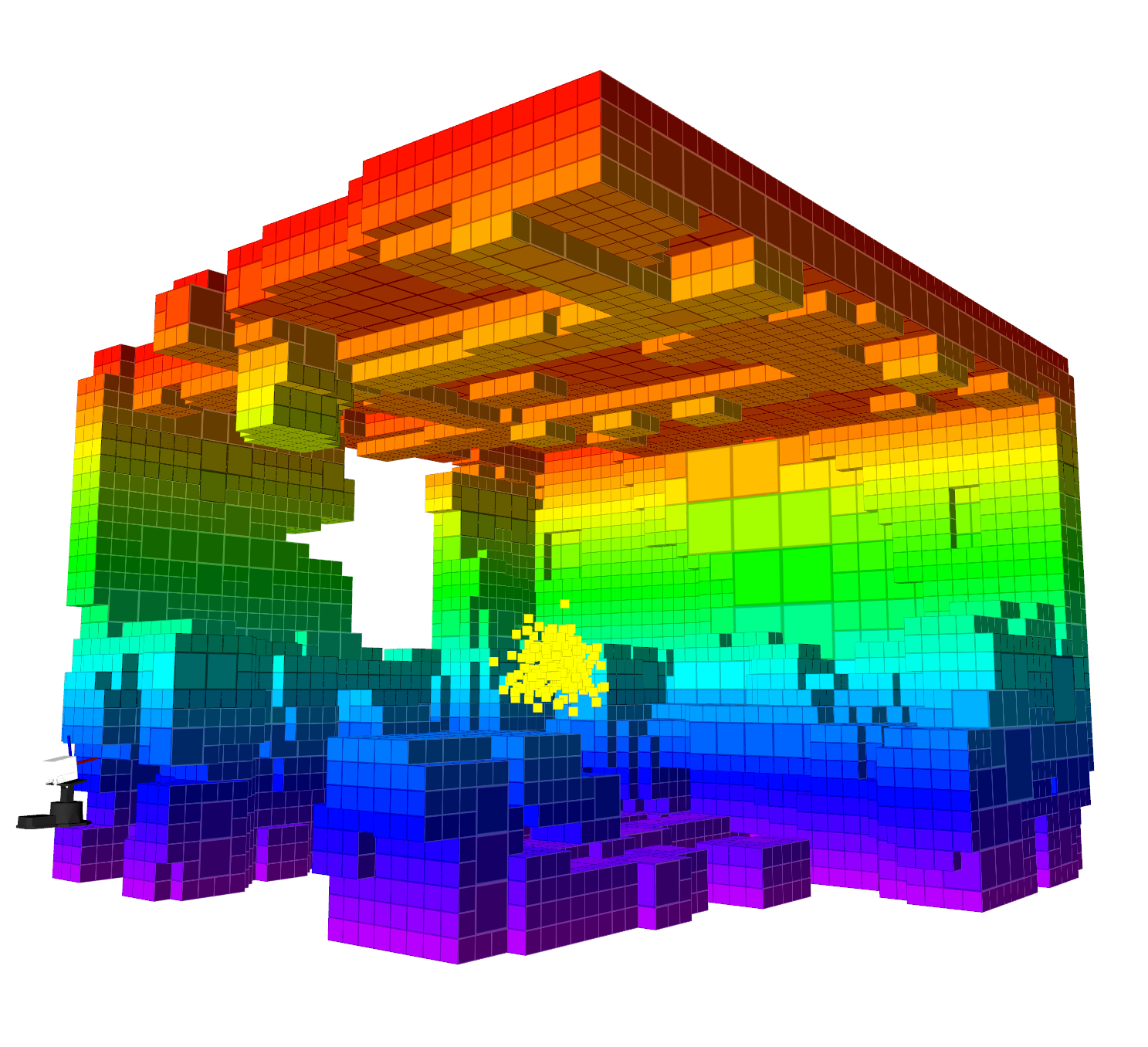}
    \caption{Stable after regain}
    \label{subfig:lost&found4}
\end{subfigure}
\caption{Different stages of the particle cloud during tracking loss and regain. The particle cloud can be seen with yellow in front of the OctoMap of the background model. The number of particles was 500 in this experiment.}
\label{fig:lost&found}
\end{figure}

\begin{figure}[t]
\centering
\includegraphics[width=8cm]{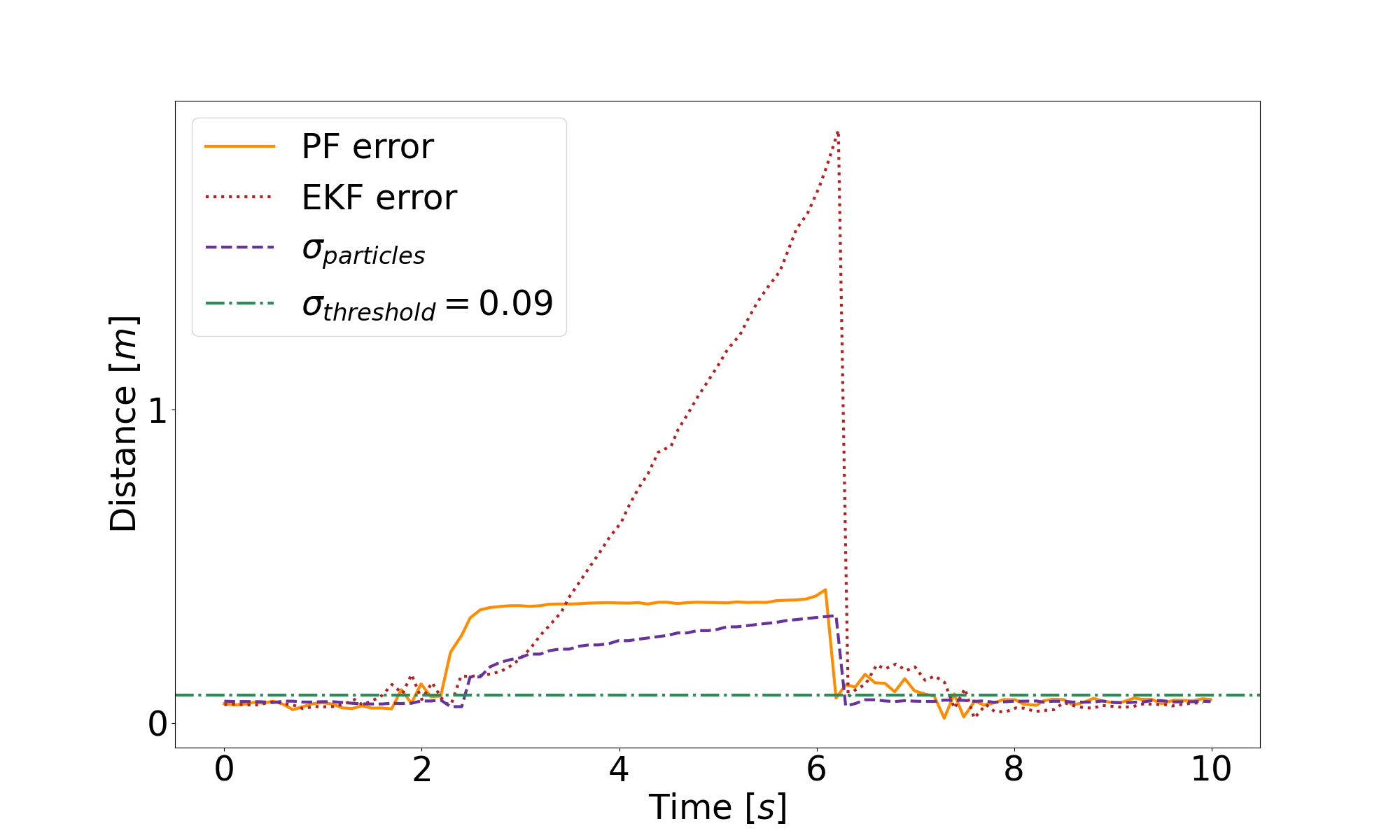}
\caption{Particle filter error and $\sigma_{particles}$ are plotted together for a typical loss and regain event. The error of the EKF of \cite{Dogru_2022} is also shown for comparison. The tracking was lost at around 2 seconds and was regained at 6 seconds.}
\label{fig:lost&found_error}
\end{figure}

FIGURE \ref{fig:lost&found} shows a typical loss and regain event of the MAV tracking from (a) to (d) while FIGURE \ref{fig:lost&found_error} shows the corresponding error and $\sigma_{particles}$ characteristics. FIGURE \ref{fig:lost&found_error} also shows the error of the EKF implementation of \cite{Dogru_2022} which varies the NNS radius based on the uncertainty, giving the method re-detection capability. This is one traversal of the drone through the 3 waypoints of the 'Lost and found' pattern. First, the normal MAV tracking is depicted in FIGURE \ref{subfig:lost&found1}. The particles are grouped around the drone with $\sigma_{particles}<\sigma_{threshold}$. When the drone starts moving, the error increases slightly but the tracking is still stable. At around 2 seconds the drone gets behind the obstacle and $\sigma_{particles}$ increases over $\sigma_{threshold}$, meaning the tracking is lost and the particle cloud starts inflating (FIGURE \ref{subfig:lost&found2}). Note on the plot that at first the $\sigma_{particles}$ jumps sharply when the $\rho$ ratio is still considerable and then settles to a much gentler slope when only the prediction noise is added to the particle positions. The particle cloud inflates further to almost $\sigma_{particles}=0.4\;m$, which is depicted in FIGURE \ref{subfig:lost&found3} while the error stays constant. This is because the drone hovers behind the obstacle, and the estimated pose is the mean of the particle positions, which is around the position where the tracking was lost. Then the drone flies out from behind the cover, and the particle update and resample steps of the filter condense the cloud instantly, which is well depicted on the sharp drop of $\sigma_{particles}$ below $\sigma_{threshold}$ in FIGURE \ref{fig:lost&found_error}. The tracking is stable again in FIGURE \ref{subfig:lost&found4}, and the error also decreases to the static error at about 6 seconds when the drone stops moving.

Overall, the regaining of tracking after loss is practically instantaneous for our PF, while the EKF recovers slightly slower. This allows robust tracking even when the trajectory is partially occluded. Notably, the PF error is constrained throughout, while the EKF increases constantly until re-detection. This difference is clearly illustrated in TABLE \ref{tab:accuracy}, where the error of the EKF of \cite{Dogru_2022} is more than twice that of the delta position PF. The error reported for \cite{catalano_2023} is less informative in this context, as their EKF implementation lacks re-detection capability. The continuously increasing error of the EKF is not preferred if the turret is moved by the filter prediction, as it may cause the turret to rotate excessively, losing the drone outside the FoV and preventing re-detection altogether.

For a real-world example where re-detection was demonstrated, kindly check out the attached video\footref{footnote:vid}.

%%%%%%%%%%%%%%%%%%%%%%%%%%%%%%%%%%%%%%%%%%%%%%%%%%%%%%%%%%%%%%%%%%%%%%%%%%%%%%%%%%%%%
%                               CONCLUSION
%%%%%%%%%%%%%%%%%%%%%%%%%%%%%%%%%%%%%%%%%%%%%%%%%%%%%%%%%%%%%%%%%%%%%%%%%%%%%%%%%%%%%

\section{CONCLUSION} \label{conclusion}

This article presents a novel tracking method utilizing a rosette scanning pattern LiDAR mounted on a pan-tilt platform. The proposed algorithm has been rigorously tested and compared to SOTA methods in both indoor and outdoor environments, demonstrating accuracy on par with the current SOTA indoor method \cite{catalano_2023} and surpassing the leading outdoor method \cite{Dogru_2022} by an $85\;\%$ increase in detection range and at least a twofold improvement in both coverage and returned point counts.

Our method now tracks all dynamic objects, and ongoing work aims to further enhance the system by differentiating between drones and other flying objects. Future research directions also include the exploration and integration of deep learning-based tracking techniques, the analysis of environmental effects like fog or rain on our algorithm, and the development of robust multi-object tracking strategies tailored for the pan-tilt configuration.

%%%%%%%%%%%%%%%%%%%%%%%%%%%%%%%%%%%%%%%%%%%%%%%%%%%%%%%%%%%%%%%%%%%%%%%%%%%%%%%%

\bibliographystyle{IEEEtran}
\bibliography{ref_v3}

\begin{thebibliography}{10}
\providecommand{\url}[1]{#1}
\csname url@samestyle\endcsname
\providecommand{\newblock}{\relax}
\providecommand{\bibinfo}[2]{#2}
\providecommand{\BIBentrySTDinterwordspacing}{\spaceskip=0pt\relax}
\providecommand{\BIBentryALTinterwordstretchfactor}{4}
\providecommand{\BIBentryALTinterwordspacing}{\spaceskip=\fontdimen2\font plus
\BIBentryALTinterwordstretchfactor\fontdimen3\font minus \fontdimen4\font\relax}
\providecommand{\BIBforeignlanguage}[2]{{%
\expandafter\ifx\csname l@#1\endcsname\relax
\typeout{** WARNING: IEEEtran.bst: No hyphenation pattern has been}%
\typeout{** loaded for the language `#1'. Using the pattern for}%
\typeout{** the default language instead.}%
\else
\language=\csname l@#1\endcsname
\fi
#2}}
\providecommand{\BIBdecl}{\relax}
\BIBdecl

\bibitem{Rejeb_2022}
A.~Rejeb, A.~Abdollahi, K.~Rejeb, and H.~Treiblmaier, ``Drones in agriculture: A review and bibliometric analysis,'' \emph{Computers and Electronics in Agriculture}, 2022.

\bibitem{Maghazei_2019}
O.~Maghazei and T.~Netland, ``Drones in manufacturing: Exploring opportunities for research and practice,'' \emph{Journal of Manufacturing Technology Management}, vol. Forthcoming, 2019.

\bibitem{Chi-2023}
N.~T.~K. Chi, L.~T. Phong, and N.~T. Hanh, ``The drone delivery services: An innovative application in an emerging economy,'' \emph{The Asian Journal of Shipping and Logistics}, 2023.

\bibitem{Shahmoradi-2020}
J.~Shahmoradi, E.~Talebi, P.~Roghanchi, and M.~Hassanalian, ``A comprehensive review of applications of drone technology in the mining industry,'' \emph{Drones}, 2020.

\bibitem{Mohsan_2023}
S.~Mohsan, N.~Othman, Y.~Li, M.~Alsharif, and M.~Khan, ``Unmanned aerial vehicles (uavs): practical aspects, applications, open challenges, security issues, and future trends,'' \emph{Intelligent Service Robotics}, 2023.

\bibitem{Wilson-2014}
R.~L. Wilson, ``Ethical issues with use of drone aircraft,'' in \emph{2014 IEEE International Symposium on Ethics in Science, Technology and Engineering}, 2014.

\bibitem{Vacca-2017}
A.~Vacca and H.~Onishi, ``Drones: military weapons, surveillance or mapping tools for environmental monitoring? the need for legal framework is required,'' \emph{Transportation Research Procedia}, 2017.

\bibitem{Lukasiewicz2022}
J.~Łukasiewicz and A.~Kobaszyńska~Twardowska, ``Proposed method for building an anti-drone system for the protection of facilities important for state security,'' \emph{Security and Defence Quarterly}, vol.~39, pp. 88--107, 2022.

\bibitem{brazeal2021risley}
R.~G. Brazeal, B.~E. Wilkinson, and H.~H. Hochmair, ``A rigorous observation model for the risley prism-based livox mid-40 lidar sensor,'' \emph{Sensors}, vol.~21, no.~14, 2021.

\bibitem{Coluccia_2020}
A.~Coluccia, G.~Parisi, and A.~Fascista, ``Detection and classification of multirotor drones in radar sensor networks: A review,'' \emph{Sensors}, 2020.

\bibitem{Zhao_2019}
J.~Zhao, X.~Fu, Z.~Yang, and F.~Xu, ``Radar-assisted uav detection and identification based on 5g in the internet of things,'' \emph{Wireless Communications and Mobile Computing}, 2019.

\bibitem{Hammer_2018}
M.~Hammer, M.~Hebel, M.~Laurenzis, and M.~Arens, ``Lidar-based detection and tracking of small uavs,'' in \emph{Emerging Imaging and Sensing Technologies for Security and Defence III; and Unmanned Sensors, Systems, and Countermeasures}, 2018.

\bibitem{Dogru_2022}
S.~Dogru and L.~Marques, ``Drone detection using sparse lidar measurements,'' \emph{IEEE Robotics and Automation Letters}, 2022.

\bibitem{vb01}
M.~Vrba and M.~Saska, ``Marker-less micro aerial vehicle detection and localization using convolutional neural networks,'' \emph{IEEE Robotics and Automation Letters}, 2020.

\bibitem{vb02}
V.~Walter, N.~Staub, A.~Franchi, and M.~Saska, ``Uvdar system for visual relative localization with application to leader-follower formations of multirotor uavs,'' \emph{IEEE Robotics and Automation Letters}, 2019.

\bibitem{vb04}
A.~Carrio, S.~Vemprala, A.~Ripoll, S.~Saripalli, and P.~Campoy, ``Drone detection using depth maps,'' in \emph{2018 IEEE/RSJ International Conference on Intelligent Robots and Systems (IROS)}, 2018.

\bibitem{Nemer_2021}
I.~Nemer, T.~Sheltami, I.~Ahmad, A.~Yasar, and M.~Abdeen, ``Rf-based uav detection and identification using hierarchical learning approach,'' \emph{Sensors}, 2021.

\bibitem{Medaiyese-2022}
O.~O. Medaiyese, M.~Ezuma, A.~P. Lauf, and I.~Guvenc, ``Wavelet transform analytics for rf-based uav detection and identification system using machine learning,'' \emph{Pervasive and Mobile Computing}, 2022.

\bibitem{Corzo_2023}
G.~Corzo, E.~Alvarez-Aros, J.~Mariño, and N.~Amézquita-Gómez, ``Military artificial intelligence applied to sustainable development projects: sound environmental scenarios,'' \emph{DYNA}, 09 2023.

\bibitem{Al-Emadi_2021}
S.~Al-Emadi, A.~Al-Ali, and A.~Al-Ali, ``Audio-based drone detection and identification using deep learning techniques with dataset enhancement through generative adversarial networks,'' \emph{Sensors}, 2021.

\bibitem{svanstrom2020realtime}
F.~Svanström, C.~Englund, and F.~Alonso-Fernandez, ``Real-time drone detection and tracking with visible, thermal and acoustic sensors,'' in \emph{2020 25th International Conference on Pattern Recognition (ICPR)}, 2021.

\bibitem{Dudczyk_2022}
J.~Dudczyk, R.~Czyba, and K.~Skrzypczyk, ``Multi-sensory data fusion in terms of uav detection in 3d space,'' \emph{Sensors}, 2022.

\bibitem{svanstrom2022}
F.~Svanström, F.~Alonso-Fernandez, and C.~Englund, ``Drone detection and tracking in real-time by fusion of different sensing modalities,'' \emph{Drones}, 2022.

\bibitem{catalano_2023}
I.~Catalano, H.~Sier, X.~Yu, T.~Westerlund, and J.~P. Queralta, ``Uav tracking with solid-state lidars: Dynamic multi-frequency scan integration,'' in \emph{2023 21st International Conference on Advanced Robotics (ICAR)}, 2023.

\bibitem{yolo2016}
J.~Redmon, S.~Divvala, R.~Girshick, and A.~Farhadi, ``You only look once: Unified, real-time object detection,'' in \emph{2016 IEEE Conference on Computer Vision and Pattern Recognition (CVPR)}, 2016.

\bibitem{MIMO_new_2022}
S.~Gazdag, A.~Kiskaroly, T.~Sziranyi, and A.~L. Majdik, ``Autonomous racing of micro air vehicles and their visual tracking within the micro aerial vehicle and motion capture (mimo) arena,'' in \emph{ISR Europe 2022; 54th International Symposium on Robotics}, 2022.

\bibitem{hammer2020imagebased}
M.~Hammer, B.~Borgmann, M.~Hebel, and M.~Arens, ``{Image-based classification of small flying objects detected in LiDAR point clouds},'' in \emph{Laser Radar Technology and Applications XXV}.\hskip 1em plus 0.5em minus 0.4em\relax International Society for Optics and Photonics, 2020.

\bibitem{Marvasti2022}
S.~M. Marvasti-Zadeh, L.~Cheng, H.~Ghanei-Yakhdan, and S.~Kasaei, ``Deep learning for visual tracking: A comprehensive survey,'' \emph{IEEE Transactions on Intelligent Transportation Systems}, 2022.

\bibitem{Adzemovic2025}
M.~Adžemović, P.~Tadić, A.~Petrović, and M.~Nikolić, ``Beyond kalman filters: deep learning-based filters for improved object tracking,'' \emph{Machine Vision and Applications}, pp. 20--42, 2025.

\bibitem{Spinello_Arras_Triebel_Siegwart_2010}
L.~Spinello, K.~Arras, R.~Triebel, and R.~Siegwart, ``A layered approach to people detection in 3d range data,'' \emph{Proceedings of the AAAI Conference on Artificial Intelligence}, 2010.

\bibitem{PointNet2016}
R.~Q. Charles, H.~Su, M.~Kaichun, and L.~J. Guibas, ``Pointnet: Deep learning on point sets for 3d classification and segmentation,'' in \emph{2017 IEEE Conference on Computer Vision and Pattern Recognition (CVPR)}, 2017.

\bibitem{PointNetPP2017}
C.~R. Qi, L.~Yi, H.~Su, and L.~J. Guibas, ``Pointnet++: Deep hierarchical feature learning on point sets in a metric space,'' in \emph{Proceedings of the 31st International Conference on Neural Information Processing Systems}, 2017.

\bibitem{VoxelNet2018}
Y.~Zhou and O.~Tuzel, ``Voxelnet: End-to-end learning for point cloud based 3d object detection,'' in \emph{2018 IEEE/CVF Conference on Computer Vision and Pattern Recognition}, 2018.

\bibitem{lang2019pointpillarsfastencodersobject}
A.~H. Lang, S.~Vora, H.~Caesar, L.~Zhou, J.~Yang, and O.~Beijbom, ``Pointpillars: Fast encoders for object detection from point clouds,'' in \emph{2019 IEEE/CVF Conference on Computer Vision and Pattern Recognition (CVPR)}, 2019.

\bibitem{DewanCaselitzTipaldi2016}
A.~Dewan, T.~Caselitz, G.~D. Tipaldi, and W.~Burgard, ``Motion-based detection and tracking in 3d lidar scans,'' in \emph{IEEE International Conference on Robotics and Automation (ICRA)}, 2016.

\bibitem{Moosmann2013ICRA}
F.~Moosmann and C.~Stiller, ``\BIBforeignlanguage{english}{Joint self-localization and tracking of generic objects in 3d range data},'' in \emph{\BIBforeignlanguage{english}{Proceedings of the {IEEE} International Conference on Robotics and Automation}}, 2013.

\bibitem{RazlawQuenzelBehnke2019}
J.~Razlaw, J.~Quenzel, and S.~Behnke, ``Detection and tracking of small objects in sparse 3d laser range data,'' in \emph{2019 International Conference on Robotics and Automation (ICRA)}, 2019.

\bibitem{AdaptiveScan2021}
L.~Qingqing, Y.~Xianjia, J.~P. Queralta, and T.~Westerlund, ``Adaptive lidar scan frame integration: Tracking known mavs in 3d point clouds,'' in \emph{2021 20th International Conference on Advanced Robotics (ICAR)}, 2021.

\bibitem{Rasshofer}
R.~Rasshofer, M.~Spies, and H.~Spies, ``Influences of weather phenomena on automotive laser radar systems,'' \emph{Advances in Radio Science}, 2011.

\bibitem{Heinzler_2019}
R.~Heinzler, P.~Schindler, J.~Seekircher, W.~Ritter, and W.~Stork, ``Weather influence and classification with automotive lidar sensors,'' in \emph{2019 IEEE Intelligent Vehicles Symposium (IV)}, 2019.

\bibitem{Bijelic_2018}
M.~Bijelic, T.~Gruber, and W.~Ritter, ``A benchmark for lidar sensors in fog: Is detection breaking down?'' in \emph{2018 IEEE Intelligent Vehicles Symposium (IV)}, 2018.

\bibitem{ParticleFilter_1993}
N.~J. Gordon, D.~J. Salmond, and A.~F. Smith, ``Novel approach to nonlinear/non-gaussian bayesian state estimation,'' in \emph{IEE proceedings F (radar and signal processing)}, 1993.

\bibitem{balla2024}
K.~Balla, A.~Keszler, S.~Gazdag, T.~Szirányi, and A.~L. Majdik, ``Detection and classification of small-sized uavs and birds in sparse lidar point cloud *,'' in \emph{2024 IEEE International Symposium on Safety Security Rescue Robotics (SSRR)}, 2024.

\bibitem{hornung13auro}
A.~Hornung, K.~M. Wurm, M.~Bennewitz, C.~Stachniss, and W.~Burgard, ``{OctoMap}: An efficient probabilistic {3D} mapping framework based on octrees,'' \emph{Autonomous Robots}, 2013.

\bibitem{Rusu_ICRA2011_PCL}
R.~B. Rusu and S.~Cousins, ``{3D is here: Point Cloud Library (PCL)},'' in \emph{{IEEE International Conference on Robotics and Automation (ICRA)}}, 2011.

\bibitem{Elfring2021}
J.~Elfring, E.~Torta, and R.~van~de Molengraft, ``Particle filters: A hands-on tutorial,'' \emph{Sensors}, 2021.

\end{thebibliography}

\end{document}